\documentclass{article}

\usepackage[preprint]{neurips_2026}

\usepackage[utf8]{inputenc} 
\usepackage[T1]{fontenc}    
\usepackage{hyperref}       
\usepackage{url}            
\usepackage{booktabs}       
\usepackage{amsfonts}       
\usepackage{nicefrac}       
\usepackage{microtype}      
\usepackage{xcolor}         
\usepackage{multirow}       
\usepackage{algorithm}
\usepackage{algpseudocode}
\usepackage{graphicx}
\usepackage{wrapfig}
\usepackage{subcaption}
\usepackage{enumitem}

\usepackage{amsmath,amssymb,amsthm}

\newtheorem{theorem}{Theorem}[section]

\newtheorem{definition}[theorem]{Definition}

\usepackage{kotex}

\title{Shaping Zero-Shot Coordination via State Blocking}

\author{%
  Mingu Kang\thanks{Equal contribution.} \quad
  Sunwoo Lee\footnotemark[1] \quad
  Yonghyeon Jo \quad
  Seungyul Han\thanks{Corresponding author: Seungyul Han.} \\
  Graduate School of Artificial Intelligence \\
  UNIST \\
  Ulsan, South Korea 44919 \\
  \texttt{\{mingukang,sunwoolee,yonghyeonjo,syhan\}@unist.ac.kr}
}

\begin{document}

\maketitle

\begin{abstract}
Zero-shot coordination (ZSC) aims to enable agents to cooperate with independently trained partners without prior interaction, a key requirement for real-world multi-agent systems and human-AI collaboration. Existing approaches have largely emphasized increasing partner diversity during training, yet such strategies often fall short of achieving reliable generalization to unseen partners. We introduce State-Blocked Coordination (SBC), a simple yet effective framework that improves ZSC by inducing diverse interaction scenarios without direct environment modification. Specifically, SBC generates a family of virtual environments through state blocking, allowing agents to experience a wide range of suboptimal partner policies. Across multiple benchmarks, SBC demonstrates superior performance in zero-shot coordination, including strong generalization to human partners.
\end{abstract}

\section{Introduction}
As multi-agent systems rapidly evolve, research has moved beyond coordination among co-trained agents toward interaction with independently trained partners, including humans. Such capabilities are critical for real-world applications, such as assistive robotics~\cite{zhao2022coordination, li2023proactive} and autonomous systems~\cite{zhang2018human, cornelisse2024human}. A key challenge in these settings is that agents must coordinate with diverse and often suboptimal human behaviors, while large-scale human interaction data remains costly and difficult to collect. This challenge has motivated Zero-Shot Coordination (ZSC)~\cite{hu2020other, strouse2021collaborating}, where an \emph{ego agent} is trained to coordinate with independently trained \emph{partner agents} without requiring access to human data.

Early ZSC methods build on Self-Play (SP)~\cite{samuel1959some, silver2018general}, where agents learn by interacting with copies of themselves. While effective in fully cooperative settings, SP often converges to a single coordination convention, resulting in poor generalization when the ego agent is paired with independently trained partners. To mitigate this limitation, prior work has explored approaches such as Other-Play~\cite{hu2020other}, which promotes symmetry-aware coordination, as well as methods that increase partner diversity through populations or randomized policies~\cite{lupu2021trajectory, strouse2021collaborating, zhao2023maximum, yan2023efficient}. However, these approaches remain limited in capturing a diverse set of structured suboptimal coordination modes, often yielding either similar behaviors or unstructured variability. More recently, environment-diversity approaches~\cite{jha2025cross} attempt to address this by training agents across varied environments, showing that robustness to environmental changes can improve generalization to unseen partners. However, such methods typically require explicit environment access, introducing additional assumptions. As a result, existing ZSC methods still face a fundamental challenge: constructing diverse yet structured suboptimal partner behaviors that enable robust generalization to unseen partners.

\begin{figure}[t!]
  \centering
  \includegraphics[width=0.8\linewidth]{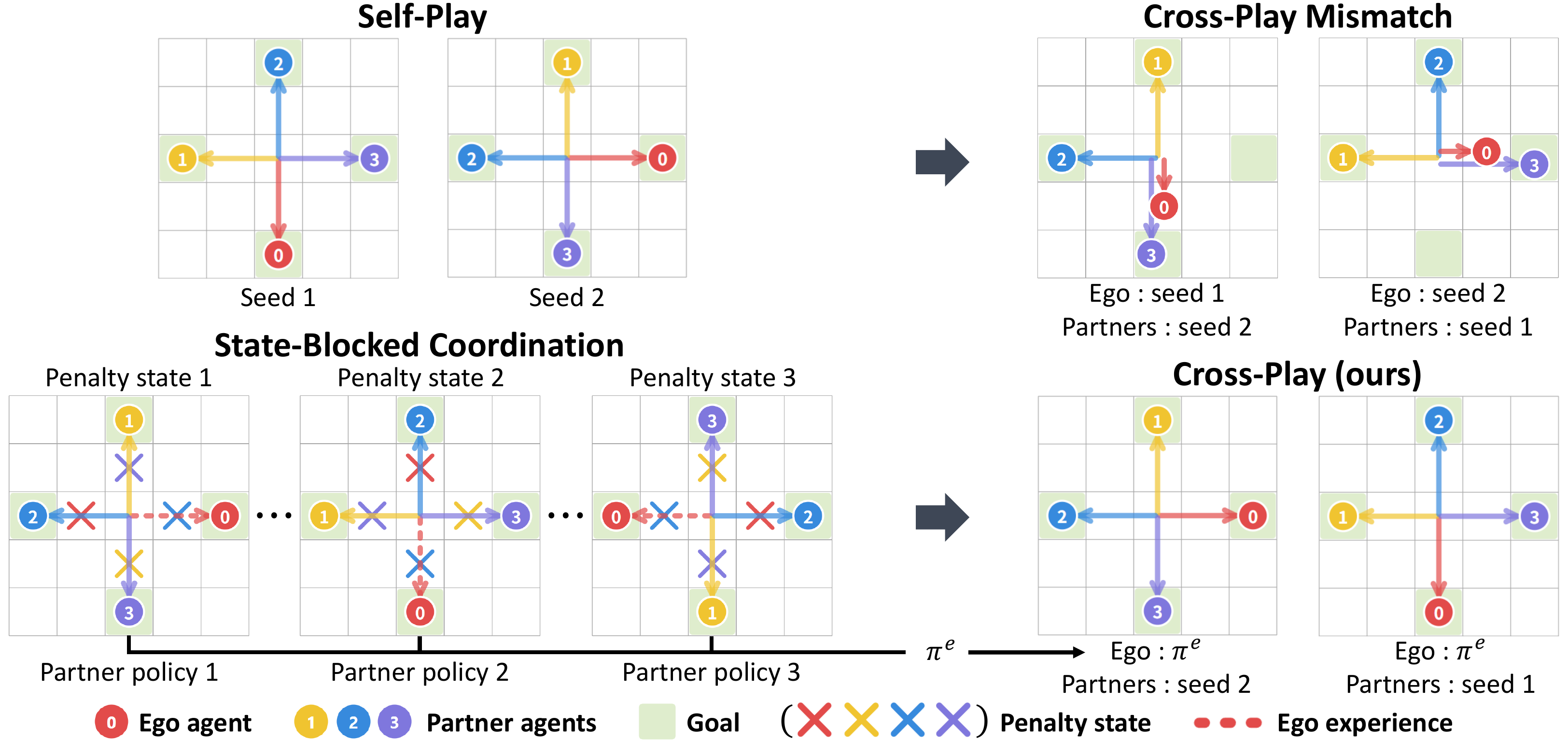}
  \vspace{-0.3em}
  \caption{Illustration of State-Blocked Coordination in Multi-Destination Spread. \textbf{Top:} Independently trained self-play policies may adopt incompatible conventions across seeds, causing failures in cross-play. \textbf{Bottom:} SBC creates virtual environmental variations through state blocking, inducing diverse suboptimal partner policies. Training the ego policy with these partners improves coordination with unseen partners following different conventions.
}
  \vspace{-0.5em}
  \label{figure1}
\end{figure}

Figure~\ref{figure1} illustrates this challenge in Multi-Destination Spread, where four agents must coordinate to reach distinct destinations for success. The upper row shows that SP-trained policies exhibit variation across random seeds, yet remain tied to similar coordination conventions, leading to failures in cross-play when independently trained agents are paired together. This mismatch highlights a key limitation of existing ZSC approaches: while they introduce some variability, they fail to produce sufficiently distinct coordination modes, resulting in limited coverage of meaningful suboptimal strategies and poor generalization to unseen partners. Consequently, a central question arises: how can we systematically expose the ego agent to a diverse set of structured suboptimal strategies without requiring explicit access to or modification of the environment?

To address this limitation, we propose \textbf{State-Blocked Coordination (SBC)}, a framework that induces structured diversity in partner behaviors through state blocking. Specifically, SBC assigns partners a designated penalty state to avoid, effectively creating a family of virtual environments without directly modifying the underlying environment. As each partner must still complete the task while avoiding its assigned state, this induces distinct, locally optimal coordination strategies corresponding to structured suboptimal behaviors. Figure~\ref{figure1} (bottom row) illustrates this effect: partners trained with different penalty states commit to different goal-reaching strategies, resulting in clearly separated coordination modes. By training the ego policy with these partners, SBC enables the ego to coordinate robustly across diverse partner conventions, leading to successful cross-play even with previously unseen partners. Beyond this illustrative example, we show that SBC consistently improves zero-shot coordination performance across multiple benchmarks, including Multi-Destination Spread and Overcooked v1, and enables effective coordination with human partners.

\section{Related Work}
\label{sec:related}

\paragraph{Zero-shot coordination methods.}
Zero-shot coordination (ZSC) trains cooperative agents that generalize to partners with independently formed conventions, and prior work mainly differs in how diversity is introduced during training. One line of work extends self-play to mitigate dependence on a single coordination convention, including symmetry-based approaches such as Other-Play~\citep{hu2020other}, equivariant architectures~\citep{muglich2022equivariant}, belief-based reasoning~\citep{hu2021off, cui2021k}, and formal analyses of convention-free coordination~\citep{treutlein2021new}. Another line focuses on exposing the ego agent to diverse partner behaviors through populations of independently trained policies or randomized mixtures that approximate such diversity~\citep{strouse2021collaborating, lupu2021trajectory, zhao2023maximum, lucas2022any, yan2023efficient}. Generative approaches further model a latent space of partner strategies and sample from it during training~\citep{derek2021adaptable, liang2024learning, li2025adaptively}. Finally, environmental diversity can be introduced by exposing agents to a broader set of environments, thereby eliciting different partner behaviors~\citep{jha2025cross}.
 \vspace{-0.5em}
\paragraph{Human--AI coordination.}
Human--AI coordination has been studied across real-world and simulated environments. In real-world settings, prior work considers domains such as human--robot collaboration~\citep{li2023proactive, zhao2022coordination}, shared autonomy~\citep{losey2020controlling}, and interactive driving~\citep{cornelisse2024human}. Complementary work evaluates coordination in cooperative games where human data can be collected at scale, most notably Hanabi~\citep{bard2020hanabi} and Overcooked~\citep{carroll2019utility}, with additional testbeds in social perception~\citep{puig2020watch}. A related line of work considers \emph{ad-hoc teamwork}~\citep{stone2010ad, 
barrett2015cooperating, chen2020aateam, rahman2021towards, wang2024n}, where the ego agent is exposed to a broader pool of partner policies, including those not encountered during training, to improve robustness to unseen teammates. In contrast, we focus on zero-shot coordination in which partner behaviors arise from the same training process, without access to externally provided or arbitrary partner policies.
 \vspace{-0.5em}
\paragraph{Reward design across RL domains.}
Modifying the reward signal to steer learning has a long history in RL, with methods differing by intent. One use is in exploration, where potential-based shaping~\citep{ng1999policy, wiewiora2003principled, devlin2012dynamic}, count-based bonuses~\citep{bellemare2016unifying, tang2017exploration, ostrovski2017count}, and prediction-error curiosity~\citep{pathak2017curiosity, burda2018exploration, burda2018large} introduce dense or novelty-driven signals to encourage state visitation. A line of work focuses on safety and constraint satisfaction, where negative rewards are applied at unsafe states to enforce constrained-MDP feasibility~\citep{altman2021constrained, ray2019benchmarking}, typically through trust-region methods~\citep{achiam2017constrained, yang2020projection} or adaptive Lagrangian penalties~\citep{tessler2018reward, stooke2020responsive}. In offline learning, conservative value penalties~\citep{kumar2020conservative, fujimoto2019off} and uncertainty-based reward penalties~\citep{yu2020mopo, kidambi2020morel} are used to discourage policies from drifting into out-of-distribution or uncertain regions. Our work adopts a similar state-dependent penalty mechanism, but uses it to steer partners toward different coordination modes rather than to enforce safety or conservatism.

\section{Background}

\paragraph{Fully cooperative Markov Decision Process.}
We consider a fully cooperative Markov decision process (MDP)~\cite{littman1994markov} defined by the tuple $\mathcal{M}=\langle \mathcal{N}, \mathcal{S}, \mathcal{A}, \mathcal{T}, \mathcal{R}, \gamma \rangle$, where $\mathcal{N}=\{1,\ldots,n\}$ denotes the set of agents, $\mathcal{S}$ the state space, and $\mathcal{A}$ the shared action space. The transition function is $\mathcal{T}: \mathcal{S}\times\mathcal{A}^n\times\mathcal{S}\rightarrow[0,1]$, and the shared reward function is $\mathcal{R}: \mathcal{S}\times\mathcal{A}^n\times\mathcal{S}\rightarrow\mathbb{R}$. At each time step $t$, each agent $i$ selects an action $a_t^i$ according to its policy $\pi^i$, inducing a joint action $\boldsymbol a_t=(a_t^1,\ldots,a_t^n)$ under the joint policy $\boldsymbol\pi(\cdot|s_t)=\prod_{i=1}^n \pi^i(\cdot|s_t)$. The environment transitions as $s_{t+1}\sim\mathcal T(\cdot|s_t,\boldsymbol a_t)$ and all agents receive a shared reward $r_t=\mathcal R(s_t,\boldsymbol a_t,s_{t+1})$. The objective is to maximize the expected discounted return $J_{\mathcal M}(\boldsymbol\pi)=\mathbb E[\sum_{t=0}^{\infty}\gamma^t r_t]$.
 \vspace{-0.5em}
\paragraph{Zero-shot coordination protocol.}
In zero-shot coordination, we consider a joint policy
$\boldsymbol{\pi}=(\pi^e,\boldsymbol{\pi}^p)$ consisting of an ego policy $\pi^e$
and partner policies $\boldsymbol{\pi}^p$, where $\pi^e$ is optimized during training and evaluated at test time.
We evaluate $\pi^e$ under AI--AI and human--AI settings.
In AI--AI evaluation, \emph{self-play (SP)} pairs $\pi^e$ with copies of itself, measuring coordination under a shared convention, while \emph{cross-play (XP)} pairs $\pi^e$ with independently trained partners from different random seeds of the same method, evaluating generalization across seed-induced conventions.
\emph{Held-out XP} further evaluates generalization by pairing $\pi^e$ with partners trained using different ZSC methods.
In human--AI evaluation, $\pi^e$ is paired with human proxies learned from demonstrations or, when available, with real human partners.

\section{Methods}
\label{sec:methods}

\subsection{State-Blocked MDP for Structured Partner Diversity}
\label{sec:sbl}

In this section, we present the formulation of \emph{State-Blocked Coordination} (SBC), which induces structured partner diversity by creating environmental variations through state blocking. As illustrated in Figure~\ref{figure1}, our key idea is to generate diverse partner behaviors by learning policies that avoid designated states, thereby inducing different suboptimal coordination modes.

Specifically, we consider a penalty-state set $\tilde{\mathcal{S}}=\{\tilde{s}_1,\tilde{s}_2,\ldots,\tilde{s}_K\}\subset\mathcal{D}$, where $\mathcal{D}$ denotes states collected during learning and $K$ is the maximum size of the set. We then construct environmental variations by treating these states as blocked states and penalizing policies that visit them, which leads to the following \emph{State-Blocked MDP} (SB-MDP).

\begin{definition}[State-Blocked Markov Decision Process]
\label{def:spmdp}
Given a cooperative MDP $\mathcal M$ and penalty-state set $\tilde{\mathcal S}$, we define a \emph{State-Blocked MDP} as $\mathcal{M}_{\tilde{\mathcal{S}}}=\langle\mathcal{N}, \mathcal{S}, \mathcal{A}, \mathcal{T}, \tilde{\mathcal{R}}, \gamma, \alpha\rangle$, where the penalized reward is
\begin{align}
\tilde{\mathcal R}(s_t,\boldsymbol a_t,s_{t+1};\tilde{\mathcal{S}}):=
\mathcal R(s_t,\boldsymbol a_t,s_{t+1}) -
\alpha\mathbb I(s_{t+1}\in\tilde{\mathcal S}),
\label{penalized_reward}
\end{align}
with penalty coefficient $\alpha\ge0$. When $K=0$, the formulation recovers the original MDP $\mathcal M$.
\end{definition}

Under the SB-MDP, visiting penalty states incurs additional penalties, encouraging policies to avoid those states and inducing locally optimal yet potentially suboptimal behaviors. As a result, the SB-MDP objective provides a lower bound on the original MDP objective:
\[
J_{\mathcal{M}_{\tilde{\mathcal{S}}}}(\boldsymbol{\pi})
\le
J_{\mathcal{M}}(\boldsymbol{\pi}),
\]
since $\tilde{\mathcal{R}}(s_t,\boldsymbol{a}_t,s_{t+1}) \le \mathcal{R}(s_t,\boldsymbol{a}_t,s_{t+1})$ for all transitions.

A key consequence is that changing the penalty-state set $\tilde{\mathcal S}$ induces different SB-MDPs $\mathcal M_{\tilde{\mathcal S}}$, each giving rise to distinct coordination modes. Thus, varying only the penalty-state set enables the construction of diverse suboptimal partner policies, which our ego policy can subsequently learn to coordinate with.
 \vspace{-0.5em}
\paragraph{Practical reward design.}
While Definition~\ref{def:spmdp} assigns penalties only when visiting states exactly in $\tilde{\mathcal S}$, in practice a smoother reward design is often preferable. We therefore consider a distance-based penalty formulation for SB-MDP,
\begin{align}
\tilde{\mathcal R}(s_t,\boldsymbol a_t,s_{t+1};\tilde{\mathcal{S}}) = 
\mathcal R(s_t,\boldsymbol a_t,s_{t+1}) -
\alpha
\sum_{\tilde s\in\tilde{\mathcal S}}
\frac{1}{\|s_{t+1}-\tilde s\|_2+\epsilon},
\label{eq:penalty}
\end{align}
where $\epsilon>0$ is a small constant for numerical stability. Under this design, states closer to penalty states incur larger penalties, yielding smoother optimization while admitting a similar proof intuition.

\subsection{State-Blocked Coordination through Blocking-aware Partner Exposure}
\label{sec:train}

We now describe the training procedure of \emph{State-Blocked Coordination} (SBC), whose goal is to expose the ego policy to diverse partner behaviors induced by different state-blocked MDPs. Since changing the penalty-state set $\tilde{\mathcal S}$ changes the optimal policy in the corresponding SB-MDP $\mathcal M_{\tilde{\mathcal S}}$, the policy must be aware of which states are blocked to solve the state-blocked task. We therefore introduce a \emph{blocking-aware} joint policy $\tilde{\boldsymbol{\pi}}=(\tilde{\pi}^e(\cdot|s_t,\tilde{\mathcal S}),\tilde{\boldsymbol{\pi}}^p(\cdot|s_t,\tilde{\mathcal S}))$, where the blocking-aware policies condition on the penalty-state set $\tilde{\mathcal S}\sim\mathcal D$, with $\mathcal D$ the replay buffer collected during training. The blocking-aware policy is trained in the SB-MDP by maximizing the penalized return
\begin{equation}
J_{\mathcal{M}_{\tilde{\mathcal S}}}(\tilde{\pi}^e)
=
\mathbb E_{\sim(\tilde{\pi}^e,\tilde{\boldsymbol{\pi}}^p)}
\left[
\sum_{t=0}^{\infty}\gamma^t\tilde r_t
\right],
\label{eq:sbobj}
\end{equation}
where $\tilde{\boldsymbol{\pi}}^p$ is instantiated as copies of $\tilde{\pi}^e$ under self-play. By sampling diverse penalty-state sets during training, SBC induces a family of blocking-aware policies solving different SB-MDPs, thereby producing structured suboptimal partner behaviors.

The second stage trains the normal ego policy $\pi^e$ in the original MDP $\mathcal M$ by exposing it to both self-play and blocking-aware partners, without providing penalty-state information as input. Let $\boldsymbol{\pi}=(\pi^e,\boldsymbol{\pi}^p)$ denote the normal joint policy. We optimize
\begin{equation}
J_{\mathcal M}(\pi^e)
=
\mathbb E_{\boldsymbol{\pi}'\sim
\mathrm{Unif}\{\boldsymbol{\pi}^p,\tilde{\boldsymbol{\pi}}^p\},\,\tilde{\mathcal S}\sim\mathcal D}
\left[
\mathbb E_{\sim(\pi^e,\boldsymbol{\pi}')}
\left[
\sum_{t=0}^{\infty}\gamma^t r_t
\right]
\right],
\label{eq:normalobj}
\end{equation}
which trains $\pi^e$ to coordinate with a broader set of structured partner behaviors rather than overfitting to a single self-play convention. Here, $\boldsymbol{\pi}^p$ is also instantiated as copies of $\pi^e$, and $\tilde{\pi}^e$ and $\pi^e$ are trained with equal iteration ratios. Importantly, while $\tilde{\pi}^e$ serves only to induce diverse partner behaviors during training, the normal ego policy $\pi^e$ is learned in the original MDP and solely used for all test-time evaluations.

\subsection{Value-Guided Penalty-State Scheduling}
\label{sec:psf}

\begin{figure}[t!]
  \centering
  \includegraphics[width=0.8\linewidth]{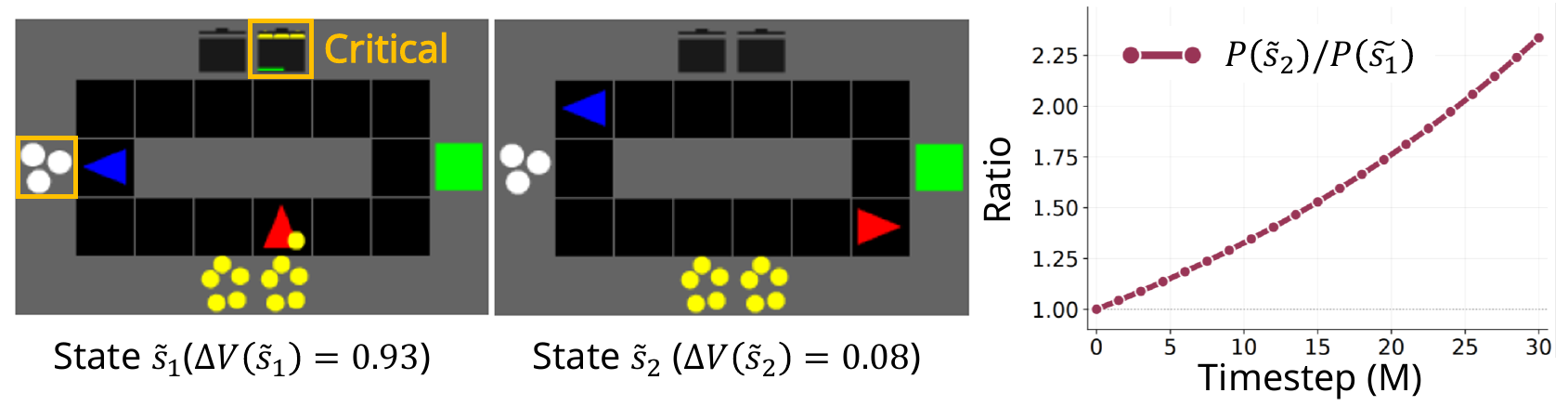}
  \vspace{-0.2cm}
  \caption{
  Visualization of value-guided penalty-state scheduling in Overcooked. \textbf{Left:} $\tilde{s}_1$ is a task-essential state, such as picking up a plate or ingredient, whose blocking induces a large value gap, whereas $\tilde{s}_2$ is a less critical state with a smaller value gap. \textbf{Right:} sampling ratio under the proposed scheduling, which are initially uniform but progressively down-weight $\tilde{s}_1$ over training. 
}
  \label{fig:value_based_sampling}
  \vspace{-0.1cm}
\end{figure}

A key component of SBC is how to select penalty states for constructing SB-MDPs, since different penalty states induce different blocking-aware partner policies. While uniform sampling maximizes diversity, blindly blocking task-essential states can induce overly degraded partner behaviors and harm the ego policy trained to coordinate with them. To address this, we propose \emph{value-guided penalty-state scheduling}, which biases sampling away from overly critical states. Let $V^{\boldsymbol{\pi}}(s_t)$ denote the value under the original MDP and $\tilde{V}^{\tilde{\boldsymbol{\pi}}}(s_t,\tilde{\mathcal S})$ the value under an SB-MDP. We quantify the criticality of a penalty state $\tilde s$ through the value gap
\begin{equation}
\Delta V(\tilde s)
:=
V^{\boldsymbol{\pi}}(s_0)
-
\tilde{V}^{\tilde{\boldsymbol{\pi}}}(s_0,\tilde s),
\label{eq:value_gap}
\end{equation}
where larger $\Delta V(\tilde s)$ indicates a more task-essential state. We then sample penalty states according to
\begin{equation}
P(\tilde s)
\propto
\exp(-\beta \Delta V(\tilde s)),
\label{eq:set_sampling}
\end{equation}
so states causing large value degradation are sampled less frequently. The temperature $\beta$ is annealed from $0$ to $1$, yielding nearly uniform sampling early in training for diversity, and progressively shifting toward safer penalty states. We construct $\tilde{\mathcal S}$ by sampling up to $K$ states from the replay buffer $\mathcal D$ according to $P(\tilde s)$, with the set size uniformly drawn from $\{1,\ldots,K\}$. Figure~\ref{fig:value_based_sampling} illustrates this intuition in Overcooked, where task-essential states induce larger value gaps and are progressively down-weighted over training. Thus, the scheduler initially explores diverse blocking configurations while gradually avoiding overly disruptive ones.

\begin{figure}[t!]
  \centering
  \includegraphics[width=0.9\linewidth]{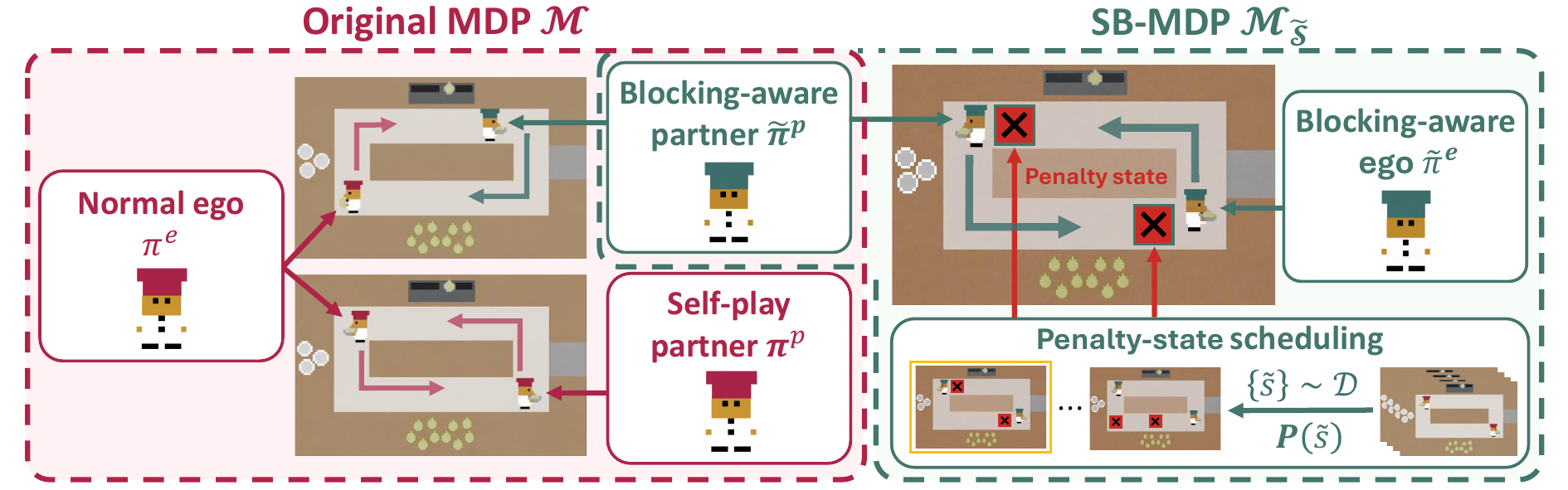}
  \vspace{-0.2cm}
  \caption{Overview of the proposed SBC framework. Value-guided scheduling selects penalty states to construct SB-MDPs for learning blocking-aware policies, while the normal ego policy is trained in the original MDP through exposure to both self-play and blocking-aware partners.
  }
  \vspace{-0.3cm}
  \label{fig:framework}
\end{figure}

To improve partner adaptation and diversity, we additionally employ partner-action prediction and random action injection following~\citep{yan2023efficient}. Figure~\ref{fig:framework} summarizes the full SBC pipeline, and Algorithm~\ref{alg:sbc} provides the complete training procedure. In our implementation, both blocking-aware and normal policies are optimized using Eqs.~\eqref{eq:sbobj} and \eqref{eq:normalobj}, instantiated with IPPO~\cite{de2020independent}, where each agent independently updates its own policy. Additional implementation details are deferred to Appendix~\ref{app:impl}.

\begin{algorithm}[h]
\caption{State-Blocked Coordination (SBC)}
\label{alg:sbc}
\begin{algorithmic}[1]
\State Initialize the normal policy $\boldsymbol{\pi}$, the blocking-aware policy $\tilde{\boldsymbol{\pi}}$, the replay buffer $\mathcal{D}$, and  $\Delta V \leftarrow 0$
\For{each training iteration}
    \State Sample a penalty state set $\tilde{\mathcal{S}} = \{\tilde{s}_1,\cdots,\tilde{s}_k\}$ from $\mathcal{D}$, where $k \sim \mathrm{Unif}(1,\cdots,K)$ \Comment{Eq.~\eqref{eq:set_sampling}}
    \State Construct the SB-MDP $\mathcal{M}_{\tilde{\mathcal{S}}}$ with the penalized reward $\tilde{\mathcal{R}}$ \Comment{Eq.~\eqref{eq:penalty}}
    \Statex \hspace{\algorithmicindent}\textbf{Blocking-aware policy training}
        \State \hspace{\algorithmicindent} Roll out $(\tilde{\pi}^e,\tilde{\boldsymbol{\pi}}^p)$ in $\mathcal{M}_{\tilde{\mathcal{S}}}$
        \State \hspace{\algorithmicindent} Update the blocking-aware ego policy $\tilde{\pi}^e$ using Eq.~\eqref{eq:sbobj}
    \Statex \hspace{\algorithmicindent}\textbf{Normal policy training}
        \State \hspace{\algorithmicindent} Sample a partner policy $\boldsymbol{\pi}' \sim \mathrm{Unif}\{\boldsymbol{\pi}^p,\, \tilde{\boldsymbol{\pi}}^p\}$ and roll out $(\pi^e, \boldsymbol{\pi}')$ in $\mathcal{M}$
        \State \hspace{\algorithmicindent} Update the normal ego policy $\pi^e$ using Eq.~\eqref{eq:normalobj}
    \State Update the value gap $\Delta V$ and store transitions in $\mathcal{D}$
\EndFor
\State \textbf{Return} the normal ego policy $\pi^e$
\end{algorithmic}
\end{algorithm}


%
 
\section{Experiments}
\label{sec:experiments}

\subsection{Experimental Setup}
\label{sec:envs}

\begin{figure}[t!]
  \centering
  \includegraphics[width=0.8\linewidth]{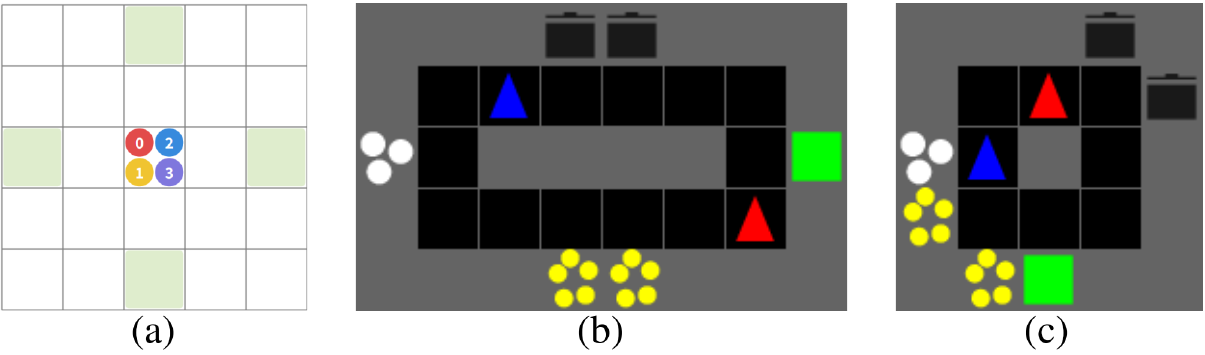}
  \caption{
Visualization of the evaluation environments.
(a) Multi-Destination Spread.
(b) \texttt{Counter Circuit} and (c) \texttt{Coordination Ring} in Overcooked v1.}
  \vspace{-0.4cm}
  \label{fig:environments1}
\end{figure}

\paragraph{Environments.}
In this section, we evaluate SBC on two environments: \textbf{Multi-Destination Spread} and \textbf{Overcooked v1}~\citep{carroll2019utility}.
As shown in Figure~\ref{fig:environments1}(a), Multi-Destination Spread is a fully observable four-agent cooperative task on a $5 \times 5$ grid, where agents start from a common central cell and aim to reach four distinct goals. The reward is proportional to the number of agents that reach different goals, with a maximum of 10 when all four agents succeed.
Overcooked v1 is a fully observable two-agent cooperative task in which agents collaborate to prepare and deliver dishes, receiving rewards upon every successful task completion.
We evaluate on five standard layouts: \texttt{Counter Circuit}, \texttt{Cramped Room}, \texttt{Coordination Ring}, \texttt{Forced Coordination}, and \texttt{Asymmetric Advantages}. Figure~\ref{fig:environments1}(b) shows example layouts.
Detailed task descriptions, state and action spaces, and reward settings are provided in Appendix~\ref{app:envs}.

\vspace{-.5em}
\paragraph{ZSC baselines.}
We compare SBC against a diverse set of ZSC baselines:
\textbf{IPPO}~\citep{de2020independent}, a self-play baseline with independent PPO agents;
\textbf{FCP}~\citep{strouse2021collaborating}, which uses a checkpoint- and seed-based partner population;
\textbf{MEP}~\citep{zhao2023maximum}, which encourages diversity by maximizing behavioral entropy;
\textbf{E3T}~\citep{yan2023efficient}, which uses mixtures of random policies as partners;
\textbf{GAMMA}~\citep{liang2024learning}, which samples partners from a VAE-based latent space;
and \textbf{CEC}~\citep{jha2025cross}, which induces partner diversity through procedurally modified layouts.
\textbf{SBC (Ours)} trains the ego agent using state-blocked partners.
For fair comparison, we follow the official implementations of prior methods. In particular, GAMMA is trained for 100M steps and CEC for 3000M (3B) steps, while the remaining baselines are trained for 30M steps. SBC is also trained for 30M steps, with hyperparameters selected via environment-specific tuning.

For ZSC evaluation, each method is trained with 10 random seeds. In the main paper, we report AI--AI performance using \textbf{SP}, \textbf{XP}, and their difference \textbf{Gap}, defined as $|\text{SP} - \text{XP}|$. Detailed seed-wise XP heatmaps are provided in Appendix~\ref{app:heatmaps}, and additional results on \textbf{Held-out XP} are reported in Appendix~\ref{app:xp_heldout}. For human--AI evaluation, we report performance with both \textbf{human proxies} and \textbf{real humans}. We report the mean and standard deviation across seeds. Detailed descriptions of the baselines and additional experimental details are provided in Appendix~\ref{app:experimental}.

\subsection{Performance on Multi-Destination Spread}
\label{sec:results_toy}
\begin{wraptable}{r}{0.55\textwidth}
  \vspace{-1.3em}
  \centering
  \caption{Episode return on Multi-Destination Spread.}
  \label{tab:gridspread_masked}
  \small
  \begin{tabular}{l r@{\,$\pm$\,}l r@{\,$\pm$\,}l r}
    \toprule
    \textbf{Method} & \multicolumn{2}{c}{\textbf{SP}} & \multicolumn{2}{c}{\textbf{XP}} & \textbf{Gap} \\
    \midrule
    IPPO   & $\mathbf{990.0}$ & $\phantom{0}0.0$ & $\phantom{0}77.4$ & $\phantom{0}7.7$ & $912.6$ \\
    FCP    & $987.0$ & $\phantom{0}2.0$ & $236.3$ & $20.1$ & $750.7$ \\
    MEP    & $960.0$ & $\phantom{0}4.6$ & $127.6$ & $15.9$ & $832.4$ \\
    GAMMA  & $942.5$ & $19.4$ & $146.1$ & $42.7$ & $796.4$ \\
    E3T    & $\mathbf{990.0}$ & $\phantom{0}0.0$ & $130.8$ & $16.1$ & $859.2$ \\
    \midrule
    SBC   & $982.0$ & $\phantom{0}7.8$ & $\mathbf{957.2}$ & $\mathbf{19.3}$ & $\mathbf{\phantom{0}24.8}$ \\
    \bottomrule
  \end{tabular}
  \vspace{-1em}
\end{wraptable}

Table~\ref{tab:gridspread_masked} reports performance on Multi-Destination Spread. We exclude CEC from comparison, as it requires explicit environment modification to induce diversity. SBC is the only method that achieves strong performance in both \textbf{SP} and \textbf{XP}, whereas all other baselines attain high SP performance but fail in XP coordination. This gap arises because, while self-play is relatively easy, the number of valid coordination modes grows factorially ($\sim 4!$), making generalization across partner policies induced by different random seeds challenging. Consequently, existing ZSC methods fail to cover the full range of coordination modes. In contrast, as illustrated in Figure~\ref{figure1}, SBC constructs diverse SB-MDPs via penalty states and trains against the resulting blocking-aware partners, exposing the ego policy to a wide range of structured behaviors. This leads to robust generalization and strong XP performance. Results on held-out XP in Appendix~\ref{app:xp_heldout} further confirm that SBC significantly outperforms prior methods.

\begin{table}[t!]
  \centering
  \small
    \caption{%
        Episode return on Overcooked v1. Values are reported as mean $\pm$ standard deviation, with \textbf{bold} indicating the best results.
    }
  \label{tab:overcooked_v1}
  \resizebox{\textwidth}{!}{%
  \setlength{\tabcolsep}{3pt}
  \begin{tabular}{l
    r@{$\;\pm\;$}l r@{$\;\pm\;$}l r
    @{\quad}
    r@{$\;\pm\;$}l r@{$\;\pm\;$}l r
    @{\quad}
    r@{$\;\pm\;$}l r@{$\;\pm\;$}l r
    @{\quad}
    r@{$\;\pm\;$}l r@{$\;\pm\;$}l r}
    \toprule
    & \multicolumn{5}{c}{\textbf{Counter Circuit}}
    & \multicolumn{5}{c}{\textbf{Coordination Ring}}
    & \multicolumn{5}{c}{\textbf{Cramped Room}}
    & \multicolumn{5}{c}{\textbf{Forced Coordination}} \\
    \cmidrule(lr){2-6} \cmidrule(lr){7-11} \cmidrule(lr){12-16} \cmidrule(lr){17-21}
    \textbf{Method}
      & \multicolumn{2}{c}{SP} & \multicolumn{2}{c}{XP} & $\text{Gap}$
      & \multicolumn{2}{c}{SP} & \multicolumn{2}{c}{XP} & $\text{Gap}$
      & \multicolumn{2}{c}{SP} & \multicolumn{2}{c}{XP} & $\text{Gap}$
      & \multicolumn{2}{c}{SP} & \multicolumn{2}{c}{XP} & $\text{Gap}$ \\
    \midrule
    IPPO
      & $158$          & $27$          & $\phantom{0}32$ & $12$          & $126$
      & $287$          & $23$          & $\phantom{0}41$ & $11$          & $246$
      & $253$          & $\phantom{0}9$ & $187$          & $36$          & $\phantom{0}66$
      & $\mathbf{200}$ & $\mathbf{29}$ & $\phantom{0}10$ & $11$          & $190$ \\
    FCP
      & $103$          & $43$          & $107$          & $17$          & $\phantom{00}\mathbf{4}$
      & $159$          & $29$          & $166$          & $11$          & $\phantom{00}7$
      & $224$          & $50$          & $227$          & $19$          & $\phantom{00}3$
      & $128$          & $37$          & $140$          & $\phantom{0}9$ & $\phantom{0}12$ \\
    MEP
      & $\phantom{0}87$ & $24$          & $\phantom{0}83$ & $11$          & $\phantom{00}\mathbf{4}$
      & $142$          & $28$          & $143$          & $11$          & $\phantom{00}\mathbf{1}$
      & $203$          & $17$          & $205$          & $\phantom{0}7$ & $\phantom{00}2$
      & $\phantom{0}38$ & $21$          & $\phantom{0}43$ & $10$          & $\phantom{00}\mathbf{5}$ \\
    E3T
      & $121$          & $31$          & $\phantom{0}38$ & $\phantom{0}9$ & $\phantom{0}83$
      & $181$          & $58$          & $\phantom{0}98$ & $24$          & $\phantom{0}83$
      & $234$          & $\phantom{0}6$ & $165$          & $27$          & $\phantom{0}69$
      & $176$          & $10$          & $105$          & $22$          & $\phantom{0}71$ \\
    GAMMA
      & $\phantom{0}66$ & $29$          & $\phantom{0}74$ & $14$          & $\phantom{00}8$
      & $143$          & $64$          & $142$          & $23$          & $\phantom{00}\mathbf{1}$
      & $181$          & $24$          & $182$          & $\phantom{0}7$ & $\phantom{00}\mathbf{1}$
      & $108$          & $35$          & $129$          & $11$          & $\phantom{0}21$ \\
    CEC
      & $\phantom{0}82$ & $31$          & $\phantom{0}52$ & $12$          & $\phantom{0}30$
      & $203$          & $40$          & $175$          & $16$          & $\phantom{0}28$
      & $221$          & $21$          & $219$          & $19$          & $\phantom{00}2$
      & $112$          & $26$          & $\phantom{0}81$ & $13$          & $\phantom{0}31$ \\
    \midrule
    \textbf{SBC}
      & $\mathbf{246}$ & $\mathbf{38}$ & $\mathbf{168}$ & $\mathbf{32}$ & $\phantom{0}78$
      & $\mathbf{333}$ & $\mathbf{13}$ & $\mathbf{316}$ & $\mathbf{18}$ & $\phantom{0}17$
      & $\mathbf{257}$ & $\phantom{0}\mathbf{5}$  & $\mathbf{256}$ & $\phantom{0}\mathbf{1}$  & $\phantom{00}\mathbf{1}$
      & $193$          & $10$          & $\mathbf{160}$ & $\mathbf{23}$ & $\phantom{0}33$ \\
    \bottomrule
  \end{tabular}%
  }
\end{table}

\begin{figure}[t!]
\vspace{-0.1cm}
  \centering
  \includegraphics[width=0.9\linewidth]{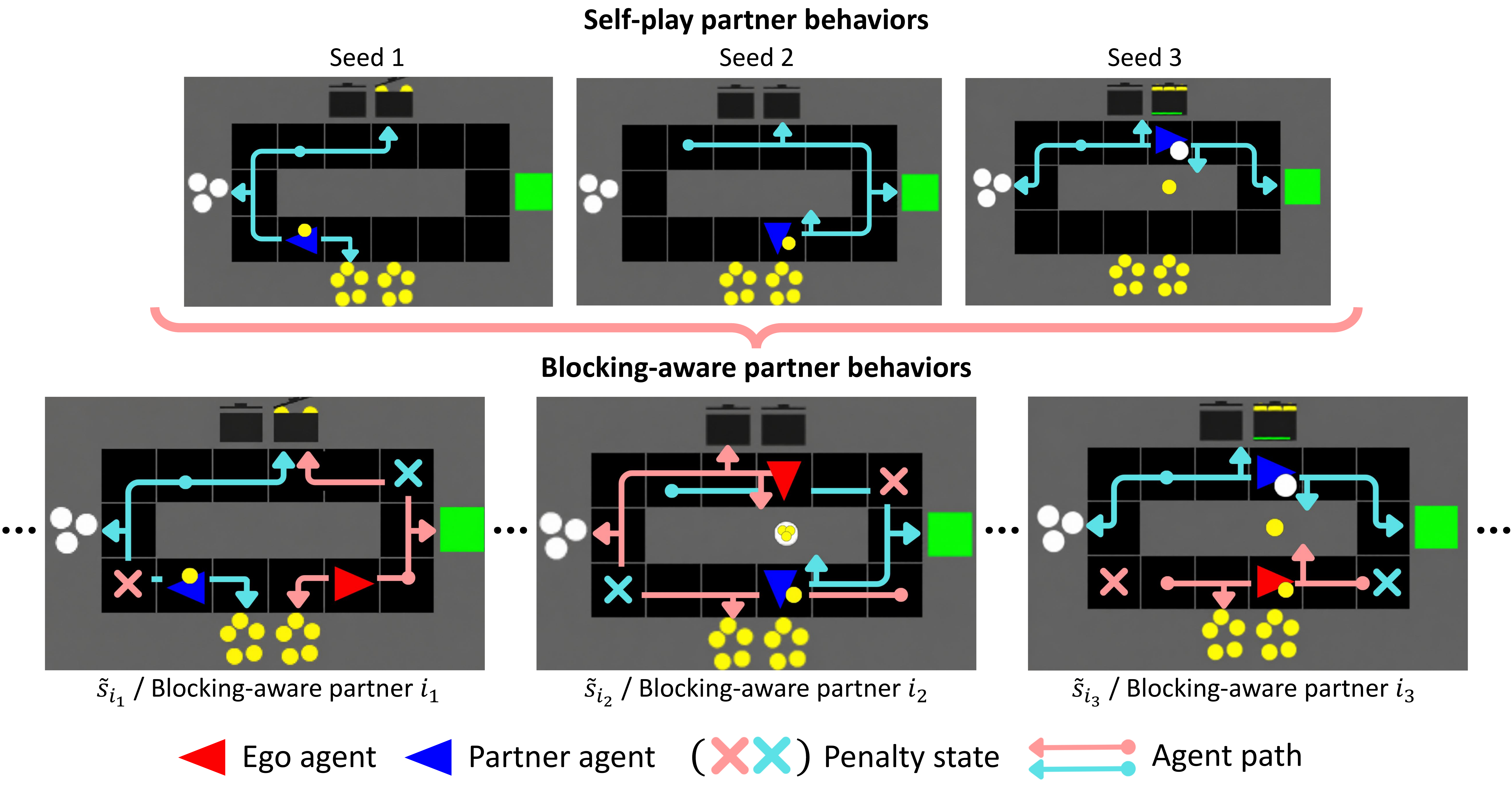}
  \vspace{-0.1cm}
  \caption{
Policy behavior analysis in Overcooked v1.
\textbf{Top:} SP partners from different seeds exhibit diverse behaviors.
\textbf{Bottom:} Blocking-aware partners induced by diverse penalty states cover this diversity, enabling robust coordination with unseen partners.
}
  \label{fig:behavior_analysis_overcooked}
  \vspace{-0.3cm}
\end{figure}

\subsection{Performance and Analysis on Overcooked v1}
\label{sec:results_overcooked}

Table~\ref{tab:overcooked_v1} reports performance on four diagnostic Overcooked v1 layouts, excluding \texttt{Asymmetric Advantages}, which yields generally high XP performance across methods. Full results including this layout, along with learning curves, are provided in Appendix~\ref{app:learning_curves}. Overall, SBC maintains strong performance in both \textbf{SP} and \textbf{XP} across most layouts. Notably, in environments such as \texttt{Counter Circuit} and \texttt{Coordination Ring}, SBC also improves SP performance over prior methods, suggesting that exposure to diverse suboptimal partners enhances overall coordination quality. Despite using fewer training samples than GAMMA and CEC, SBC achieves substantial gains while avoiding explicit environment modification for inducing partner diversity. Consistent with these results, SBC also achieves the best performance in held-out XP in Appendix~\ref{app:xp_heldout}, demonstrating robust generalization to unseen partners from other ZSC methods.

To better understand these improvements, Figure~\ref{fig:behavior_analysis_overcooked} illustrates how SBC generalizes to diverse partner behaviors induced by SP in Overcooked v1. Under SP (top row), different random seeds lead to a wide range of partner behaviors, as reflected in diverse movement patterns. SBC addresses this variability by constructing SB-MDPs with penalty states that primarily alter agent movement paths while preserving key task-critical states (e.g., ingredients and dishes). This induces diverse blocking patterns that encourage agents to learn alternative coordination strategies by navigating around blocked regions (bottom row). As a result, SBC generates a set of partner behaviors that effectively covers the variations observed under SP, enabling the ego policy to robustly adapt to partners from different seeds and achieve strong XP performance. This provides insight into why SBC achieves superior performance. Additional trajectory analyses on other Overcooked v1 layouts are provided in Appendix~\ref{app:traj_full}, showing similar trends.

\subsection{Ablation Analysis on Overcooked v1}
\label{sec:ablation}

In this section, we analyze the impact of key components and hyperparameters of SBC. In the main paper, we focus on results on \texttt{Counter Circuit}, where XP performance differences are most pronounced. Additional results on other environments, along with a comparison of computational cost, which shows that SBC incurs only moderate overhead while remaining substantially more efficient than the considered baselines, are provided in Appendix~\ref{app:analysis}.

For component ablation, we consider the following variants: \textbf{SBC (Ours)}; \textbf{SBC w/o action prediction}, which removes partner action prediction; \textbf{SBC w/ strict penalty}, which applies penalties only upon exact visits as in Eq.~\eqref{penalized_reward}; \textbf{SBC w/ uniform sampling}, which samples penalty states uniformly without value guidance; \textbf{SBC w/o SP partner}; and \textbf{SBC w/o blocking-aware partner}, which trains the ego policy with only one type of partner. As shown in Table~\ref{tab:component_ablation}, all components contribute to XP performance. Removing action prediction leads to only minor degradation, while strict penalties and uniform sampling significantly degrade performance, highlighting the importance of distance-based penalties and value-guided state sampling. Moreover, both SP and blocking-aware partners are essential, and removing either results in substantial performance drops.

We further analyze the effect of key hyperparameters in Figure~\ref{fig:hyperparam_ablation}, focusing on (a) the penalty coefficient $\alpha$ and (b) the maximum penalty-state set size $K$. We evaluate $\alpha \in \{0, 0.05, 0.1, 0.5\}$ and $K \in \{0,1,2,3,4\}$. When $\alpha=0$, state blocking is disabled, resulting in the worst performance. A moderate value ($\alpha=0.1$) achieves the best balance, while overly small or large values degrade performance due to insufficient or excessive penalization. Similarly, $K=0$ (no penalty) yields the lowest performance, and the best results are obtained at $K=1$, whereas larger values degrade performance by overly restricting the state space. Additional ablations across all layouts are provided in Appendix~\ref{app:hyperparam_full}. While $K=1$ performs best in most cases, larger values can be beneficial in environments with many optimal coordination modes under SP; for example, \texttt{Coordination Ring} achieves the best performance at $K=2$.

\begin{table}[t!]
  \centering
  \raisebox{0.40cm}{%
  \begin{minipage}[b]{0.40\linewidth}
    \scriptsize
    \centering
    \resizebox{\linewidth}{!}{%
    \begin{tabular}{l r@{\,$\pm$\,}l}
      \toprule
      \textbf{Variant}
        & \multicolumn{2}{c}{\textbf{XP}} \\
      \midrule
      SBC (Ours)
        & $\mathbf{168.0}$ & $32.0$ \\
       w/o action prediction
        & $141.7$ & $28.6$ \\
       w/ strict penalty
        & $122.9$ & $25.3$ \\
       w/ uniform sampling
        & $119.2$ & $30.4$ \\
       w/o SP partner
        & $\phantom{0}68.6$ & $19.7$ \\
       w/o blocking-aware partner
        & $\phantom{0}38.3$ & $\phantom{0}8.5$ \\
      \bottomrule
    \end{tabular}
    }
    \vspace{0.5cm}
    \caption{Component ablation.}
    \label{tab:component_ablation}
  \end{minipage}%
  }%
  \hfill
  \begin{minipage}[b]{0.58\linewidth}
    \centering
    \includegraphics[width=\linewidth]{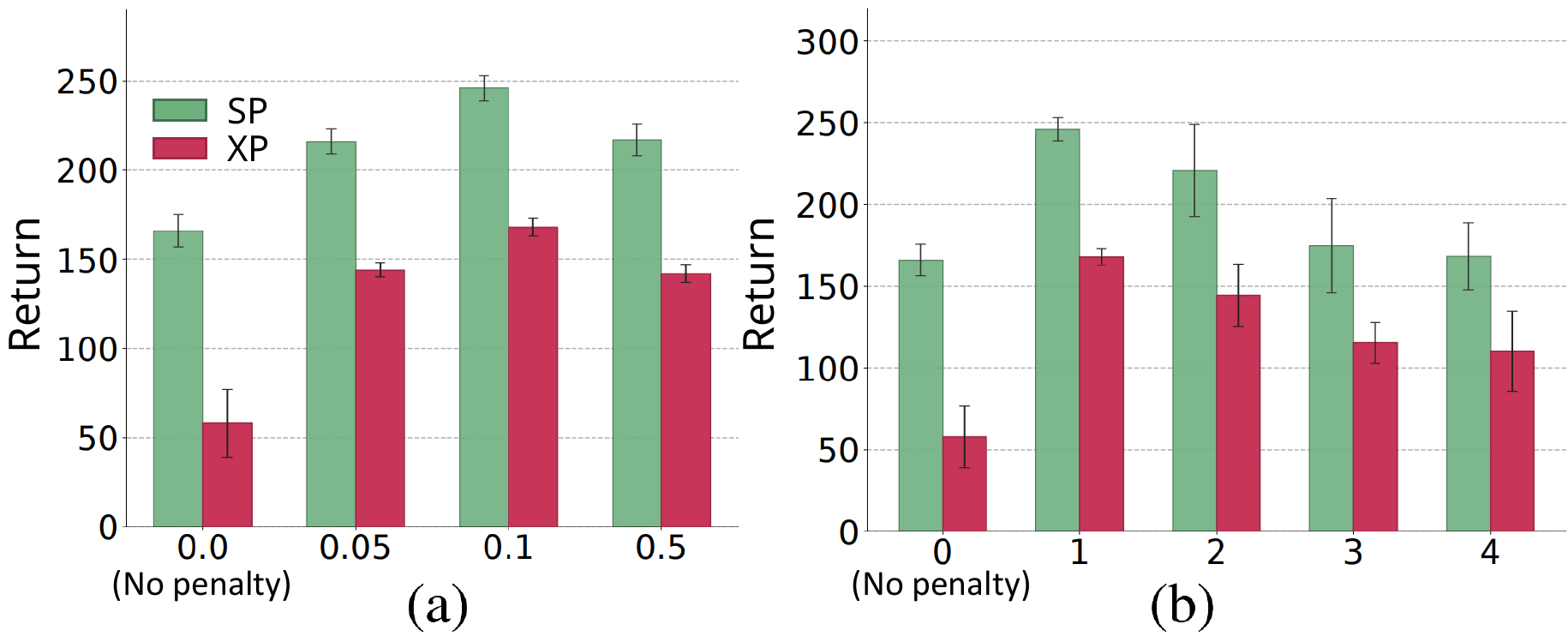}
    \vspace{-0.2cm}
    \captionof{figure}{%
      Hyperparameter analysis.
      \textbf{(a)} Penalty coefficient $\alpha$ and 
      \textbf{(b)} maximum size of penalty-state set $K$.
    }
    \label{fig:hyperparam_ablation}
  \end{minipage}
  \vspace{-2em}
\end{table}

\subsection{Zero-Shot Human--AI Coordination}
\label{sec:human_eval}

\begin{figure}[t!]
  \centering
  \includegraphics[width=0.85\linewidth]{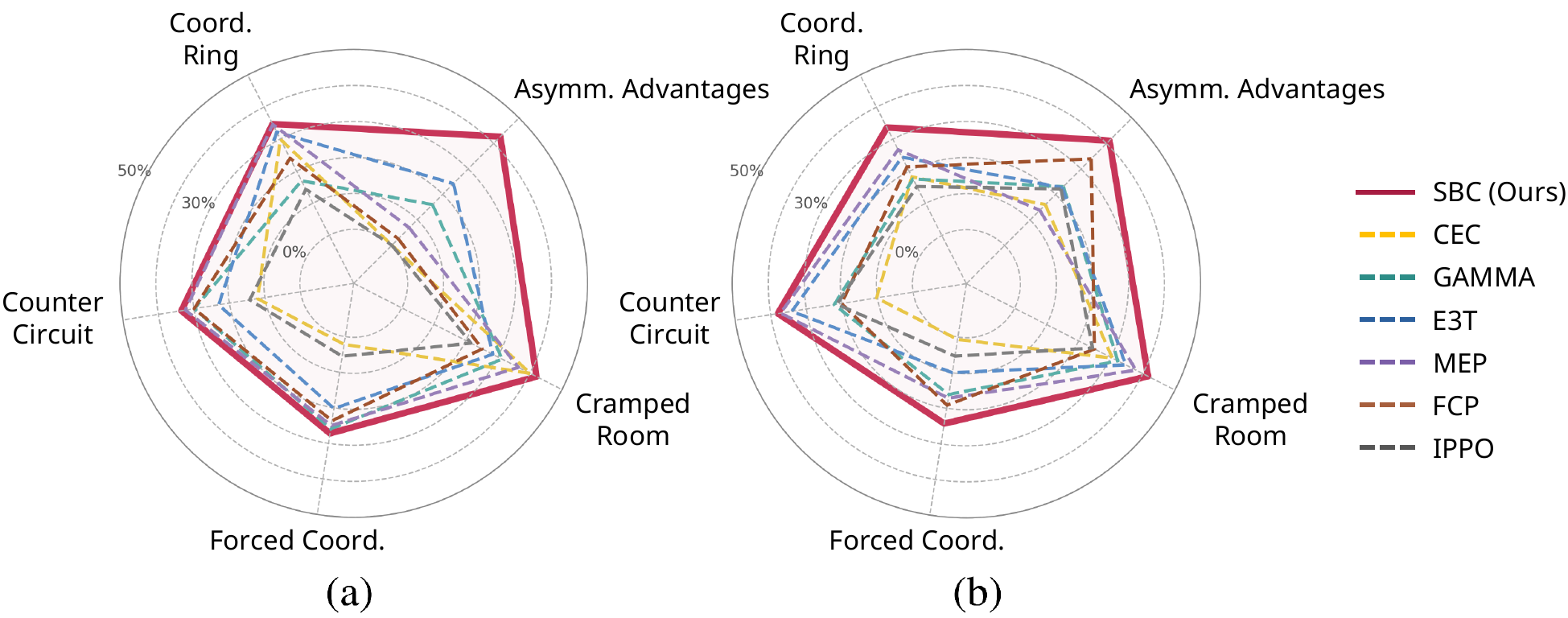}
  \caption{%
    Human--AI evaluation on Overcooked v1 with scores normalized by each layout's IPPO SP score.
\textbf{(a)} Evaluation with human proxies.
\textbf{(b)} Evaluation with real humans (\(p\text{-value} < 0.001\)).
  }
  \label{fig:human_eval}
\end{figure}

To evaluate the transfer of SBC to human--AI coordination, we conduct experiments in the Overcooked v1 environment using two protocols: \textbf{human proxies} trained on human demonstrations~\citep{carroll2019utility}, and \textbf{real humans} under an IRB-approved protocol.
In the real-human study, 40 participants played 60-second episodes with ego agents
sampled from five random seeds per method. Method identity was kept hidden, and the presentation order was counterbalanced. All participants provided informed consent and were compensated for their time.
Figure~\ref{fig:human_eval} reports scores on Overcooked v1 normalized by each
layout's IPPO SP score, where human--AI scores are generally lower than AI--AI scores due to variation in participants' Overcooked experience. Overall, SBC achieves the highest performance on all five layouts under both protocols,
whereas prior methods perform well only on specific layouts. 
These results show that exposure to diverse suboptimal blocking-aware partners is
also effective for human--AI coordination, enabling SBC to coordinate robustly
with human partners.

\begin{figure}[t!]
  \centering
  \includegraphics[width=0.75\linewidth]{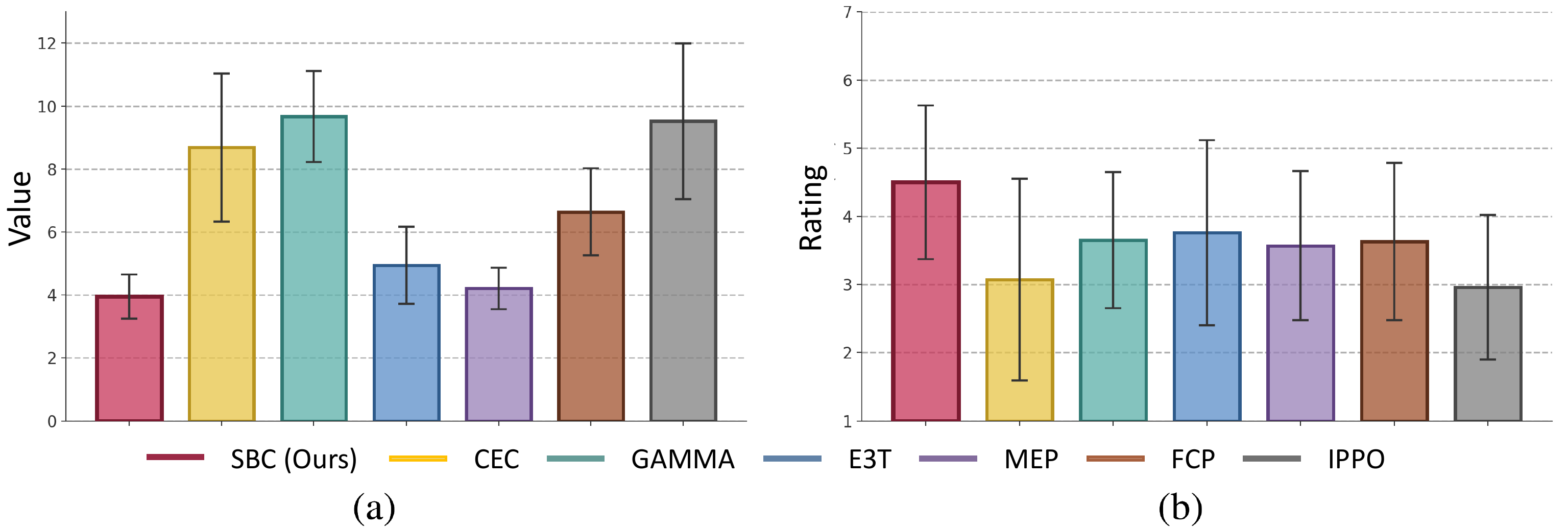}
\caption{%
    Human--AI behavioral analysis on Overcooked v1, averaged over five layouts
    (Wilcoxon signed-rank, \(p\text{-value} < 0.01\)).
    \textbf{(a)} Collisions per delivery (lower is better).
    \textbf{(b)} Participant-rated coordination fluency on a 7-point Likert scale (higher is better).
  }
  \vspace{-0.3cm}
  \label{fig:human_analysis}
\end{figure}

We further use real-human episodes to compute two behavioral metrics, averaged across the five Overcooked v1 layouts. \textbf{Collisions} quantify the number of collisions between the AI and human agents per delivery, serving as an objective measure of coordination failure. \textbf{Coordination fluency} is assessed via a post-episode 7-point Likert scale (1\,=\,very poor, 7\,=\,very good), capturing participants’ perceived smoothness of collaboration.
Figure~\ref{fig:human_analysis}(a) shows that SBC achieves the lowest number of collisions per delivery, outperforming all baselines, which reflects efficient coordination by accounting for both task completion and physical conflicts. In addition, Figure~\ref{fig:human_analysis}(b) shows that SBC receives the highest ratings on the 7-point Likert scale.
Together, these results indicate that SBC not only improves episode return but
also flexibly adapts its coordination convention to diverse human behaviors, reducing physical conflicts and enabling smoother, more
human-compatible collaboration.


\section{Limitations}
\label{sec:limitations}

Despite the strong performance of SBC, it introduces two additional hyperparameters, the penalty coefficient $\alpha$ and the maximum size of the penalty-state set $K$, which require tuning. Nevertheless, SBC consistently provides substantial performance gains, and a good configuration can be found with moderate variation, as supported by our hyperparameter analysis. As the number of agents increases, the number of possible coordination
conventions may also grow, leaving the use of larger values of $K$ in more
complex multi-agent settings as a direction for future work. In addition, computing the value gap introduces extra computational cost; however, as shown in Appendix~\ref{app:compute}, this overhead remains modest compared to IPPO and is reasonable given the improved efficiency over other ZSC methods.
\section{Conclusion}
\label{sec:conclusion}

We introduced State-Blocked Coordination (SBC), a framework for zero-shot coordination that generates structured partner diversity by penalizing designated states.
Unlike prior approaches that rely on population-level diversity or environment modification, SBC creates a family of blocking-aware partners that adopt distinct coordination modes, exposing the ego agent to broad coverage of meaningful locally optimal behaviors.
Experiments on Multi-Destination Spread and Overcooked v1 demonstrate that SBC achieves the highest cross-play performance while maintaining strong self-play scores.
Human--AI evaluations show that SBC successfully coordinates with unseen human partners.

\bibliographystyle{unsrt}     
\bibliography{references}





\clearpage

\appendix


\section{Implementation Details}
\label{app:impl}

This section provides additional implementation details of SBC.
We first describe penalty-state set construction, then present the RL training
implementation used to optimize the ego and blocking-aware policies, along with
additional training components such as partner-action prediction and random
action injection.

\paragraph{Penalty-state set construction.}
We compute the value gap for replay-buffer states using Eq.~\eqref{eq:value_gap}
and construct an episode-level penalty-state set by sampling up to $K$ states
according to Eq.~\eqref{eq:set_sampling}. The sampled penalty states are concatenated with the original state and used as
inputs to the blocking-aware policy and value function in the SB-MDP. Since the input reserves $K$ penalty-state slots, unused slots
are padded with the dummy value $-1$. In Overcooked, penalty states are
represented by the CNN-encoded state features, matching the representation used
by the policy.

\paragraph{SBC Training Details.}
We optimize both the blocking-aware ego policy $\tilde{\pi}^e$ and the normal ego
policy $\pi^e$ using Independent PPO (IPPO)~\citep{de2020independent}.
The blocking-aware policies are trained in the SB-MDP
$\mathcal{M}_{\tilde{\mathcal{S}}}$ with the penalized reward $\tilde{r}_t$, as
in Eq.~\eqref{eq:sbobj}, while the normal ego policy is trained in the original
MDP $\mathcal{M}$ with the original reward $r_t$, as in
Eq.~\eqref{eq:normalobj}.

The value estimates for the blocking-aware and normal objectives are computed by
separate critics, $\tilde{V}_{\tilde{\theta}}^{\tilde{\boldsymbol{\pi}}}$ and
$V_{\theta}^{\boldsymbol{\pi}}$, parameterized by $\tilde{\theta}$ and $\theta$,
respectively. The value losses are
\begin{align*}
    \mathcal{L}_{V}^{\mathrm{SB}}(\tilde{\theta})
    &=
    \mathbb{E}_{t}
    \left[
    \left(
    \tilde{V}_{\tilde{\theta}}^{\tilde{\boldsymbol{\pi}}}
    (s_t,\tilde{\mathcal{S}})
    -
    \hat{G}_t
    \right)^2
    \right], \\
    \mathcal{L}_{V}(\theta)
    &=
    \mathbb{E}_{t}
    \left[
    \left(
    V_{\theta}^{\boldsymbol{\pi}}(s_t)
    -
    G_t
    \right)^2
    \right],
\end{align*}
where $\tilde{G}_t$ and $G_t$ are GAE-based return targets computed
using the penalized reward $\tilde{r}_t$ and the original reward $r_t$, respectively.

Both policies are updated with the clipped IPPO objective:
\begin{align*}
    \mathcal{L}_{\mathrm{IPPO}}
    =
    \mathbb{E}_{t}
    \left[
    \min
    \left(
    \frac{\pi}{\pi_{\mathrm{old}}}\hat{A}_t,\,
    \mathrm{clip}
    \left(
    \frac{\pi}{\pi_{\mathrm{old}}},
    1-\varepsilon,1+\varepsilon
    \right)
    \hat{A}_t
    \right)
    \right],
\end{align*}
where $\pi$ denotes either $\pi^e$ or $\tilde{\pi}^e$, parameterized by $\phi$
and $\tilde{\phi}$, respectively.
The advantage $\hat{A}_t$ is computed with the corresponding reward and critic:
$(r_t,V_\theta^{\boldsymbol{\pi}})$ for $\pi^e$ and
$(\tilde{r}_t,\tilde{V}_{\tilde{\theta}}^{\tilde{\boldsymbol{\pi}}})$ for $\tilde{\pi}^e$.
Here, $\varepsilon$ is the PPO clipping coefficient.

\paragraph{Additional Training Components.}
To further improve ZSC generalization, we incorporate partner-action prediction
and random action injection into SBC, following E3T~\citep{yan2023efficient}.
For partner-action prediction, we add an auxiliary model $f_{\mathrm{pred}}$ that
predicts the partner joint action from the current state,
$\hat{\boldsymbol{a}}_t^p=f_{\mathrm{pred}}(\cdot\mid s_t)$.
It is trained with a supervised cross-entropy loss over partner actions observed
during rollouts:
\begin{align*}
    \mathcal{L}_{\mathrm{pred}}
    =
    \mathbb{E}
    \left[
    -\log f_{\mathrm{pred}}(\boldsymbol{a}_t^p\mid s_t)
    \right],
\end{align*}
where $\boldsymbol{a}_t^p$ is the partner joint action at time $t$.
The predicted action is concatenated with the ego policy input, yielding
$\pi^e(a_t^e\mid s_t,\hat{\boldsymbol{a}}_t^p)$.

For random action injection, let $\boldsymbol{\pi}^r$ be a joint random policy over
partner actions and let $\mu\in[0,1]$ be the mixing ratio.
We inject random actions into both the normal and blocking-aware partner policies:
\begin{align*}
    \boldsymbol{\pi}_{\mu}^p
    =
    (1-\mu)\boldsymbol{\pi}^p
    +
    \mu\boldsymbol{\pi}^r,
    \qquad
    \tilde{\boldsymbol{\pi}}_{\mu}^p
    =
    (1-\mu)\tilde{\boldsymbol{\pi}}^p
    +
    \mu\boldsymbol{\pi}^r.
\end{align*}
Thus, partners occasionally follow random actions during both normal ego-policy
training and blocking-aware ego-policy training, improving robustness to partner
deviations.

\newpage

\renewcommand{\theequation}{B.\arabic{equation}}
\setcounter{equation}{0}
\renewcommand{\thefigure}{B.\arabic{figure}}
\setcounter{figure}{0}
\renewcommand{\thetable}{B.\arabic{table}}
\setcounter{table}{0}
\section{Environment Details}
\label{app:envs}
In this section, we provide additional details on the benchmark environments used in our experiments. Specifically, Appendix~\ref{app:env_multi} introduces Multi-Destination Spread, where the goal is to maximize coverage by having agents reach distinct destinations, whereas Appendix~\ref{app:env_ovc1}, describes  Overcooked v1~\citep{carroll2019utility}, which aims to collaboratively complete cooking and delivery tasks.

\subsection{Multi-Destination Spread}
\label{app:env_multi}

 \begin{figure}[h!]
  \centering
  \includegraphics[width=0.6\linewidth]{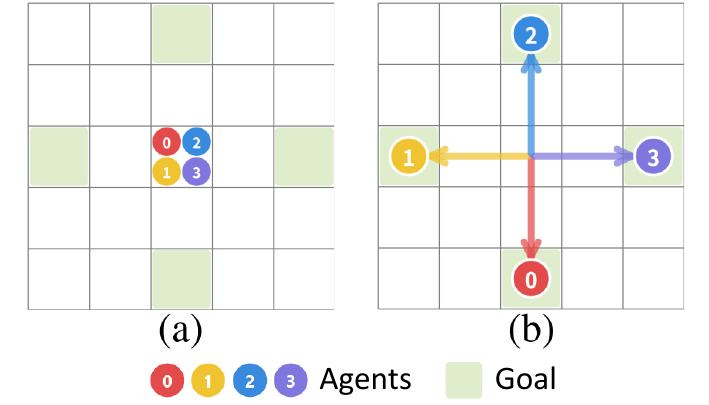}
  \caption{
Multi-Destination Spread.
(a) Four agents spawn at the same central position.
(b) Agents succeed by reaching four distinct goals.
}
  \label{fig:environments}
\end{figure}

\label{app:env_multi}

Multi-Destination Spread is a fully observable cooperative grid-world task with
four agents on a $5\times5$ grid.
All agents spawn at the center of the grid, and four goals are placed around the
center in different directions.
The objective is for the agents to cover four distinct goals at the same time.

\paragraph{State space.}
The environment state consists of all agent positions, all fixed goal positions,
the current timestep, and the terminal flag.
Each agent receives a fully observable vector observation composed of an agent
identity one-hot vector and the global state.
Specifically, the observation is
\[
    o_t^i =
    \left[
    \mathrm{onehot}(i),\,
    x_t^1,y_t^1,\ldots,x_t^4,y_t^4,\,
    g^1_x,g^1_y,\ldots,g^4_x,g^4_y
    \right],
\]
which has dimension $20$.

\paragraph{Action space.}
Each agent has a discrete action space with 9 actions:
moving in the eight grid directions or staying in place.
If an action would move an agent outside the grid, the position is clipped to
remain within the boundary.
When multiple agents attempt to occupy the same cell, the colliding agents are
reverted to their previous positions.

\paragraph{Reward function.}
Agents share a cooperative shaped reward based on the number of distinct goals
covered.
If all four goals are covered, the environment gives a reward of $10$.
If three, two, or one goals are covered, the reward is $5$, $2$, or $1$,
respectively; otherwise it is $0$.
Episodes terminate after a maximum of 100 steps.

\newpage 
 
\subsection{Overcooked v1}
\label{app:env_ovc1}
 
\begin{figure}[h!]
  \centering

  \includegraphics[width=0.8\linewidth]{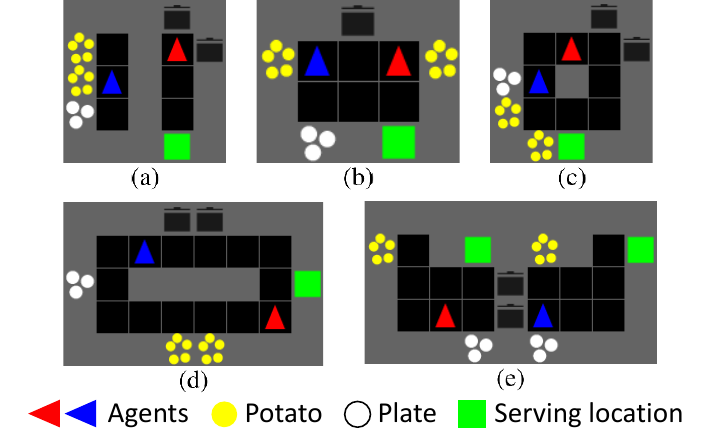}
  \caption{
Overcooked v1 layouts.
(a) \texttt{Forced Coordination}.
(b) \texttt{Cramped Room}.
(c) \texttt{Coordination Ring}.
(d) \texttt{Counter Circuit}.
(e) \texttt{Asymmetric Advantages}.
}
  \label{fig:app_env_ovc1}
\end{figure}

Overcooked v1~\citep{carroll2019utility} (MIT License) is a fully observable 
two-agent cooperative task where agents collaborate in a grid-world kitchen 
to prepare and deliver dishes.
Our implementation follows the Overcooked v1 setup provided by 
JaxMARL~\citep{flair2024jaxmarl} (Apache License 2.0).
Each layout specifies the positions of counters, pots, ingredient dispensers,
serving locations, and agent start positions.
In our experiments, we evaluate five standard layouts:
\texttt{Counter Circuit}, \texttt{Cramped Room}, \texttt{Coordination Ring},
\texttt{Forced Coordination}, and \texttt{Asymmetric Advantages}.

\paragraph{State space.}
The environment is fully observable, each agent receives a spatial tensor
observation of size $H \times W \times 26$, where $H$ and $W$ are the layout height
and width, and the 26 channels encode sparse binary features of the kitchen state.
These features include the positions and orientations of both agents, object
locations and contents such as potatoes, dishes, and soups, and the status of cooking
pots.
We process this observation with a CNN encoder followed by fully connected layers
to obtain a 128-dimensional vector representation used by the policy.

\paragraph{Action space.}
Each agent has a discrete action space with 6 actions:
moving in the four cardinal directions, interacting, or staying in place.
The interaction action allows the agent to pick up or place objects, add potatoes to
pots, pick up completed soups, and deliver dishes depending on the agent's
location and held object.

\paragraph{Reward function.}
Agents share a cooperative reward of $+20$ for each successfully delivered soup, matching the standard Overcooked-AI scale.
We also use the shaped rewards provided by JaxMARL during training: placing a potato in a pot gives $+3$, picking up a plate gives $+3$, and picking up a ready soup gives $+5$.
The distance-based shaping terms for dishes, pots, and soups are set to $0$.
Episodes are run for a fixed horizon in each layout.

\newpage 

\renewcommand{\theequation}{C.\arabic{equation}}
\setcounter{equation}{0}
\renewcommand{\thefigure}{C.\arabic{figure}}
\setcounter{figure}{0}
\renewcommand{\thetable}{C.\arabic{table}}
\setcounter{table}{0}
\section{Experimental Details}
\label{app:experimental}

All experiments are conducted on a machine with NVIDIA RTX 3090 GPUs and an
Intel Xeon Gold 6348 CPU @ 2.60\,GHz (28 cores per socket, 2 sockets, 56
physical cores), running Ubuntu 20.04 with JAX (jaxlib).
Appendix~\ref{app:hyperparameters} provides the hyperparameters used in our
experiments, Appendix~\ref{app:other_baselines} describes the ZSC baselines,
Appendix~\ref{app:zsc_eval} describes the ZSC evaluation protocol, and
Appendix~\ref{app:web_interface} presents the interface used for real-human
evaluation.

\subsection{Hyperparameters}
\label{app:hyperparameters}

All ZSC baselines and SBC follow the IPPO baseline implementation in
JaxMARL~\citep{flair2024jaxmarl}.
We use the same common hyperparameter settings as this setup, as summarized in
Table~\ref{tab:hyperparams}.
The main baseline-specific hyperparameters are provided in
Table~\ref{tab:baseline_hyperparams}.

For SBC, the penalty coefficient $\alpha$ and the maximum size of the penalty-state set $K$ are the main hyperparameters.
We select the best values for each environment through the ablation study in Appendix~\ref{app:hyperparam_full}.
The temperature $\beta$ used in value-guided penalty-state scheduling is annealed from $0$ to $1$ during training.
The environment-specific SBC hyperparameters are summarized in Table~\ref{tab:sbc_hyperparams}.

\begin{table}[h!]
  \centering
  \caption{Common IPPO hyperparameter settings shared across all methods within each environment.}
  \label{tab:hyperparams}
  \small
  \begin{tabular}{lll}
    \toprule
    \textbf{Hyperparameter} & \textbf{Multi-Destination Spread} & \textbf{Overcooked v1} \\
    \midrule
    Network architecture       & MLP (64, 1 layer) + actor/critic heads & CNN (32) + GRU (128) + FC (128) \\
    Optimizer                  & Adam ($\epsilon=10^{-5}$) & Adam ($\epsilon=10^{-5}$) \\
    Learning rate (actor / critic) & $5 \times 10^{-4}$ & $2.5 \times 10^{-4}$ \\
    Discount factor $\gamma$   & 0.99                   & 0.99 \\
    GAE $\lambda$              & 0.95                   & 0.95 \\
    PPO clip $\varepsilon$     & 0.2                    & 0.2 \\
    Entropy coef.              & 0.01                   & 0.01 \\
    Parallel envs       & 256             & 64\\
    Batch size    & 25{,}600           & 16{,}384 \\
    Buffer size             & 100               & 256 \\
    Update epochs              & 60                     & 4 \\
    \bottomrule
  \end{tabular}
\end{table}

\begin{table}[h!]
  \centering
  \caption{Baseline-specific hyperparameter settings.}
  \label{tab:baseline_hyperparams}
  \small
  \begin{tabular}{lll}
    \toprule
    \textbf{Method} & \textbf{Hyperparameter} & \textbf{Value} \\
    \midrule
    \multirow{2}{*}{FCP}
      & Population size     & 10 SP seeds $\times$ 3 checkpoints = 30 partners \\
      & Checkpoint interval & 10M timesteps (ckpt at 10M / 20M / 30M) \\
    \midrule
    MEP
      & Population size     & 4 \\
    \midrule
    \multirow{4}{*}{GAMMA}
      & VAE latent dim ($z$) & 16 \\
      & VAE hidden size & 256 \\
      & VAE KL penalty (init $\to$ final) & 0.01 $\to$ 1.0 \\
    \midrule
    E3T
      & Random-policy mixing ratio ($\epsilon$) & 0.3 \\
    \midrule
    CEC
      & Number of procedural environments & 5 (Overcooked v1 base layouts with IK) \\
    \bottomrule
  \end{tabular}
\end{table}

\begin{table}[h!]
  \centering
  \caption{Environment-specific SBC hyperparameter settings.}
  \label{tab:sbc_hyperparams}
  \small
  \begin{tabular}{llcc}
    \toprule
    \textbf{Environment} & \textbf{Layout} & \textbf{$\alpha$} & \textbf{$K$} \\
    \midrule
    Multi-Destination Spread & -- & 0.01 & 1 \\
    \midrule
    \multirow{5}{*}{Overcooked v1}
      & \texttt{Counter Circuit} & 0.1 & 1 \\
      & \texttt{Cramped Room} & 0.1 & 1 \\
      & \texttt{Coordination Ring} & 0.05 & 2 \\
      & \texttt{Forced Coordination} & 0.5 & 1 \\
      & \texttt{Asymmetric Advantages} & 0.05 & 1 \\
    \bottomrule
  \end{tabular}
\end{table}

\newpage 

\subsection{Other Baselines}
\label{app:other_baselines}
All ZSC baselines use IPPO as the base learner, following the implementation in
JaxMARL~\citep{flair2024jaxmarl}.

\paragraph{FCP.}
Fictitious Co-Play (FCP)~\citep{strouse2021collaborating} trains the ego agent
against a population of self-play agents and their historical checkpoints.
The key idea is to approximate diverse partner conventions by collecting policies
from different seeds and different stages of self-play training.
The ego policy is trained as a best response to partners sampled from this
population:
\begin{align*}
    J_{\mathrm{FCP}}(\pi^e)
    =
    \mathbb{E}_{\boldsymbol{\pi}^p \sim \mathcal{P}_{\mathrm{FCP}}}
    \left[
    J_{\mathcal{M}}(\pi^e,\boldsymbol{\pi}^p)
    \right],
\end{align*}
where $\mathcal{P}_{\mathrm{FCP}}$ denotes the partner population.

\paragraph{MEP.}
Maximum Entropy Population-based training (MEP)~\citep{zhao2023maximum}
encourages diversity by training a population of partner policies with an
additional population-entropy reward.
Let $\bar{\boldsymbol{\pi}}^p(\cdot \mid s_t)$ denote the mean policy of the
partner population.
MEP trains the partner population by maximizing the task reward together with
the entropy of the mean partner policy:
\begin{align*}
    J_{\mathrm{MEP}}(\boldsymbol{\pi}^p)
    =
    \mathbb{E}_{\boldsymbol{\pi}^p}
    \left[
    \sum_{t=0}^{T}
    \gamma^t
    \left(
    r_t
    +
    \alpha
    H(\bar{\boldsymbol{\pi}}^p(\cdot \mid s_t))
    \right)
    \right],
\end{align*}
where $\alpha$ controls the strength of the population-entropy bonus.
After training the diverse partner population, the ego agent is trained by
sampling partners from this population.

\paragraph{E3T.}
Efficient End-to-End Training (E3T)~\citep{yan2023efficient} is a ZSC method that
avoids pretraining a large partner population.
Instead, it constructs a partner policy by mixing the ego policy with a random
policy, combining coordination ability with behavioral diversity.
The partner policy is defined as
\begin{align*}
    \boldsymbol{\pi}^{p}_{\epsilon}
    =
    (1-\epsilon)\boldsymbol{\pi}^{p}
    +
    \epsilon \boldsymbol{\pi}^{r},
\end{align*}
where $\boldsymbol{\pi}^{r}$ is a random policy and $\epsilon$ is the mixing
ratio.
E3T also uses partner-action prediction, which feeds the predicted partner action
into the ego policy to help it adapt to partners following different conventions.

\paragraph{GAMMA.}
Generative Agent Modeling for Multi-agent Adaptation (GAMMA)~\citep{liang2024learning}
learns a generative model of partner behavior.
It trains a variational autoencoder (VAE) on trajectory data, where the latent
variable captures partner strategy.
After training, partners are sampled from the latent space and used to train the
ego agent against a broader distribution of generated partner strategies.

\paragraph{CEC.}
Cross-Environment Cooperation (CEC)~\citep{jha2025cross} introduces diversity by
modifying the environment rather than by constructing a partner population.
During training, the agent is trained via self-play across a distribution of
procedurally generated environments, replacing partner-population diversity
with environment diversity. This encourages the agent to learn general
coordination rules that transfer to unseen partners and unseen layouts.
We follow the official implementation provided by the
authors.\footnote{\url{https://github.com/KJha02/crossEnvCooperation}}

\newpage 

\subsection{ZSC Evaluation Protocol}
\label{app:zsc_eval}

This section describes the evaluation protocol used to assess the generalization
ability of each ZSC method to unseen AI and human partners.
We organize the evaluation into two categories: AI--AI evaluation and human--AI
evaluation.

\paragraph{AI--AI evaluation.}
AI--AI evaluation measures how well an ego policy coordinates with AI partners.
We report three metrics: self-play (SP), cross-play (XP), and
held-out XP. As explained in Section~\ref{sec:envs}, \textbf{Gap}, defined as $|\text{SP} - \text{XP}|$, measures performance degradation caused by convention mismatch. A smaller gap indicates that the ego policy is less sensitive to partner seed variation.
\begin{itemize}[leftmargin=1.3em,itemsep=0.15em,topsep=0.2em]
    \item \textbf{Self-play (SP).}
    SP evaluates coordination under the same learned convention by pairing an ego
    policy with partner policies trained from the same random seed.
    It measures task-solving performance when agents share the same training history.

    \item \textbf{Cross-play (XP).}
    XP evaluates generalization across independently learned conventions.
    For each method, we train ego policies with 10 random seeds and pair each ego
    policy with partner policies trained from different seeds of the same method.
    We report the average performance over all cross-seed pairings.

\item \textbf{Held-out XP.} Held-out XP measures whether ego policies can coordinate with partners induced by different ZSC algorithms by testing cross-method pairings. We pair each ego policy with partner policies from the various methods, including IPPO, FCP, MEP, E3T, GAMMA, and SBC. CEC was excluded due to its incompatible observation encoding.

\end{itemize}

\paragraph{Human--AI evaluation.}
Human--AI evaluation measures how well an ego policy coordinates with human-like
or real human partners on Overcooked v1 layouts.
We consider two protocols: human-proxy evaluation and real-human evaluation. We assess coordination quality using two behavioral metrics, where \textbf{Collisions} measure coordination failures by counting blocked movement attempts per delivery and \textbf{Coordination fluency} captures subjective collaboration quality via participant ratings on a seven-point Likert scale.

\begin{itemize}[leftmargin=1.3em,itemsep=0.15em,topsep=0.2em]
    \item \textbf{Human proxies.}
    Human-proxy evaluation uses a learned proxy partner to approximate human
behavior. Following the protocol in Overcooked~\citep{carroll2019utility}, we
train a partner policy by behavior cloning on human demonstration data.
We use the human demonstration data and behavior cloning model provided by the
Overcooked-AI codebase.\footnote{\url{https://github.com/HumanCompatibleAI/overcooked_ai}}
At test time, each ego policy is paired with this behavior-cloned human proxy.
This protocol allows us to evaluate whether the ego policy can coordinate with
human-like behavior without requiring real-time human participants.

    \item \textbf{Real humans.}
    Real-human evaluation pairs each ego policy with actual human participants in
real time. We conduct a study with 40 participants, following the IRB protocol
of our institution. Each participant plays 60-second episodes with ego agents
sampled from five random seeds per method. We evaluate seven methods across five Overcooked v1 layouts, resulting in 35 episodes per participant.The method identity
is hidden from participants, and the presentation order is counterbalanced across
participants.



\end{itemize}

\newpage

\subsection{Interface for Real-Human Evaluation}
\label{app:web_interface}

\begin{figure}[h]
  \centering
  \begin{subfigure}[t]{0.6\linewidth}
    \centering
    \includegraphics[width=\linewidth]{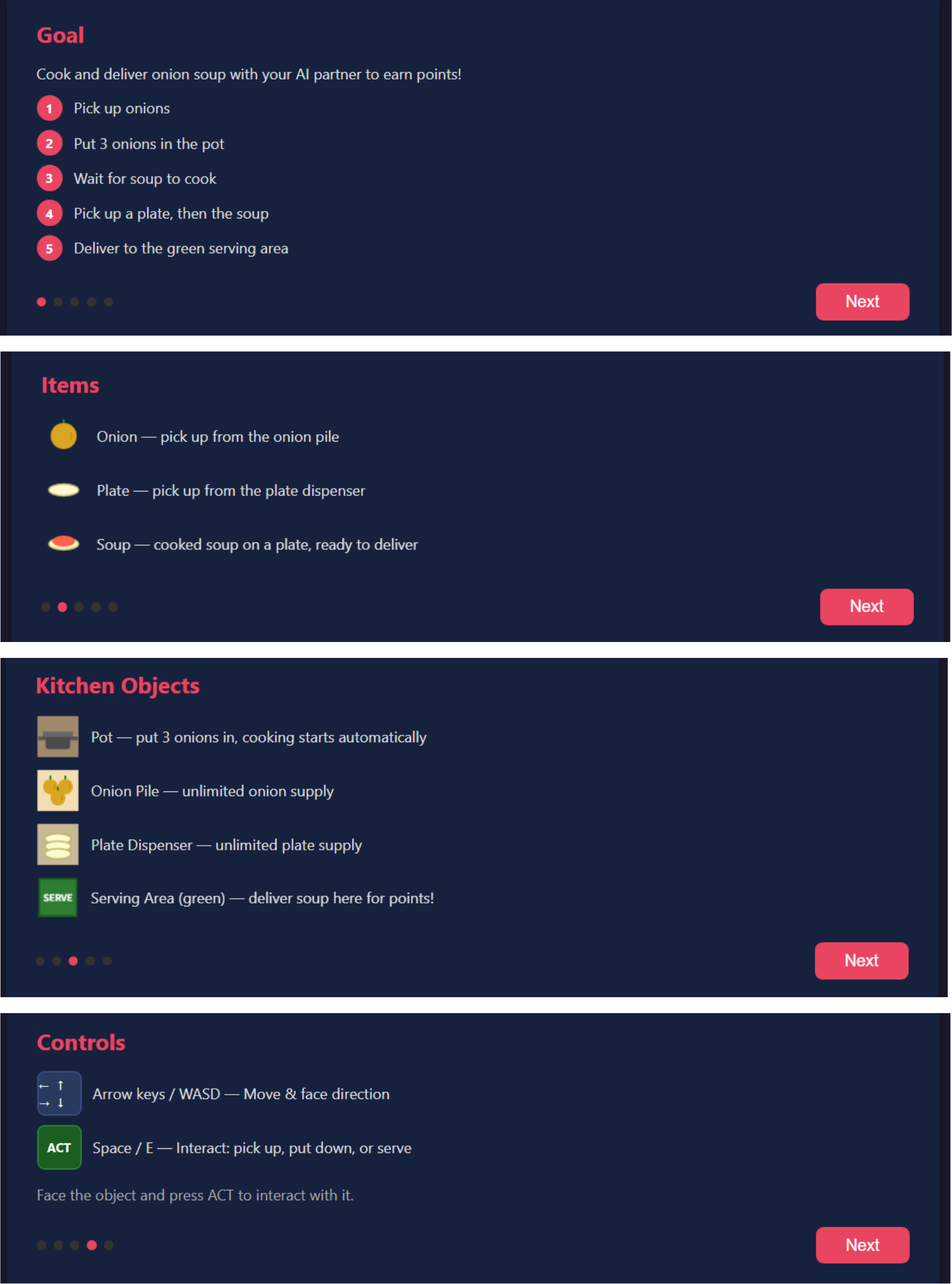}
    \caption{Pre-session instructions.}
    \label{fig:appendix_interface_instructions}
  \end{subfigure}

  \vspace{1em}

  \begin{subfigure}[t]{0.6\linewidth}
    \centering
    \includegraphics[width=\linewidth]{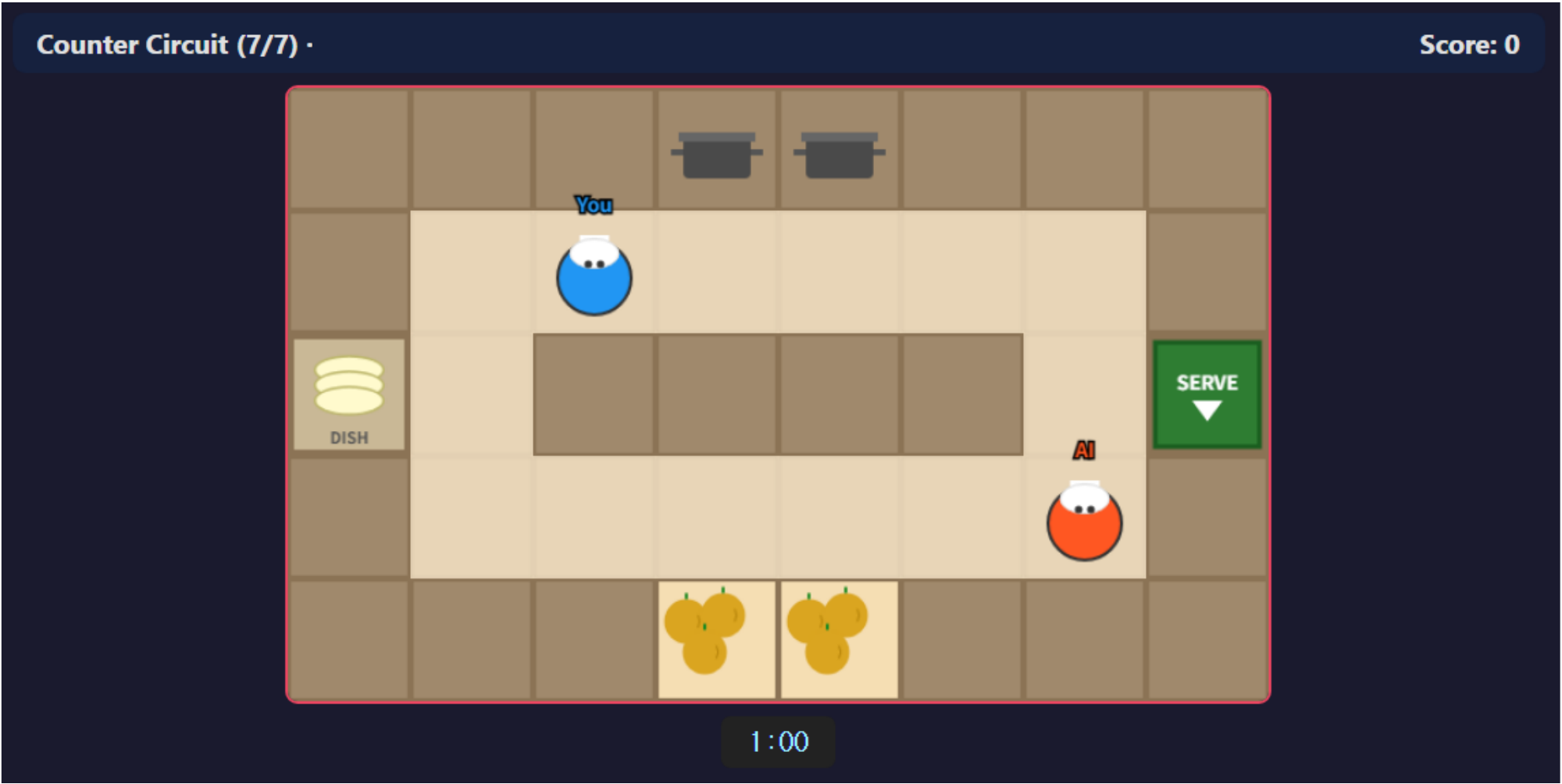}
    \caption{Gameplay screen.}
    \label{fig:appendix_interface_game}
  \end{subfigure}
  \caption{Web interface used in the real-human evaluation.
  (a)~Pre-session instructions covering the goal, items, kitchen objects,
  and controls.
  (b)~The gameplay screen, where the participant (blue, ``You'') is paired
  with an ego agent (orange, ``AI'') in a shared kitchen.}
  \label{fig:appendix_interface}
\end{figure}

To conduct the real-human evaluation in Appendix~\ref{app:zsc_eval}, we
developed a custom web-based interface that lets participants play
Overcooked v1 in real time with our trained ego agents through a standard
browser. Before the main session, each participant is guided through a
brief instruction sequence
(Figure~\ref{fig:appendix_interface_instructions}), and then plays
episodes on the gameplay screen
(Figure~\ref{fig:appendix_interface_game}). The method identity is hidden
and the presentation order is counterbalanced across participants. The
interface logs trajectories and collision events for behavioral analysis,
and after each episode participants rate the seven-point Likert item
``\emph{The AI partner and I coordinated well as a team}'' (1--7).

\newpage 

\renewcommand{\theequation}{D.\arabic{equation}}
\setcounter{equation}{0}
\renewcommand{\thefigure}{D.\arabic{figure}}
\setcounter{figure}{0}
\renewcommand{\thetable}{D.\arabic{table}}
\setcounter{table}{0}
\section{Additional Performance Results}
\label{app:perf}
This section provides additional performance comparisons between SBC and ZSC
baselines.
Appendix~\ref{app:heatmaps} presents seed-wise XP heatmaps for each method.
Appendix~\ref{app:xp_heldout} reports held-out XP performance not only across
different seeds but also across unseen methods.
Appendix~\ref{app:learning_curves} provides tables and learning curves for all layouts of Overcooked v1.

\subsection{Seed-wise XP Heatmaps}
\label{app:heatmaps}

This section presents seed-wise XP heatmaps for SBC and the ZSC baselines on
Multi-Destination Spread and Overcooked v1.
Each heatmap reports the cross-play performance obtained by pairing ego policies
and partner policies trained with different random seeds.
Rows correspond to ego seeds, columns correspond to partner seeds, and each cell
shows the mean episode reward of the corresponding ego--partner pair.

\paragraph{Multi-Destination Spread.}
\label{app:heatmap_dual}

\begin{figure}[h]
  \centering
  \includegraphics[width=0.6\linewidth]{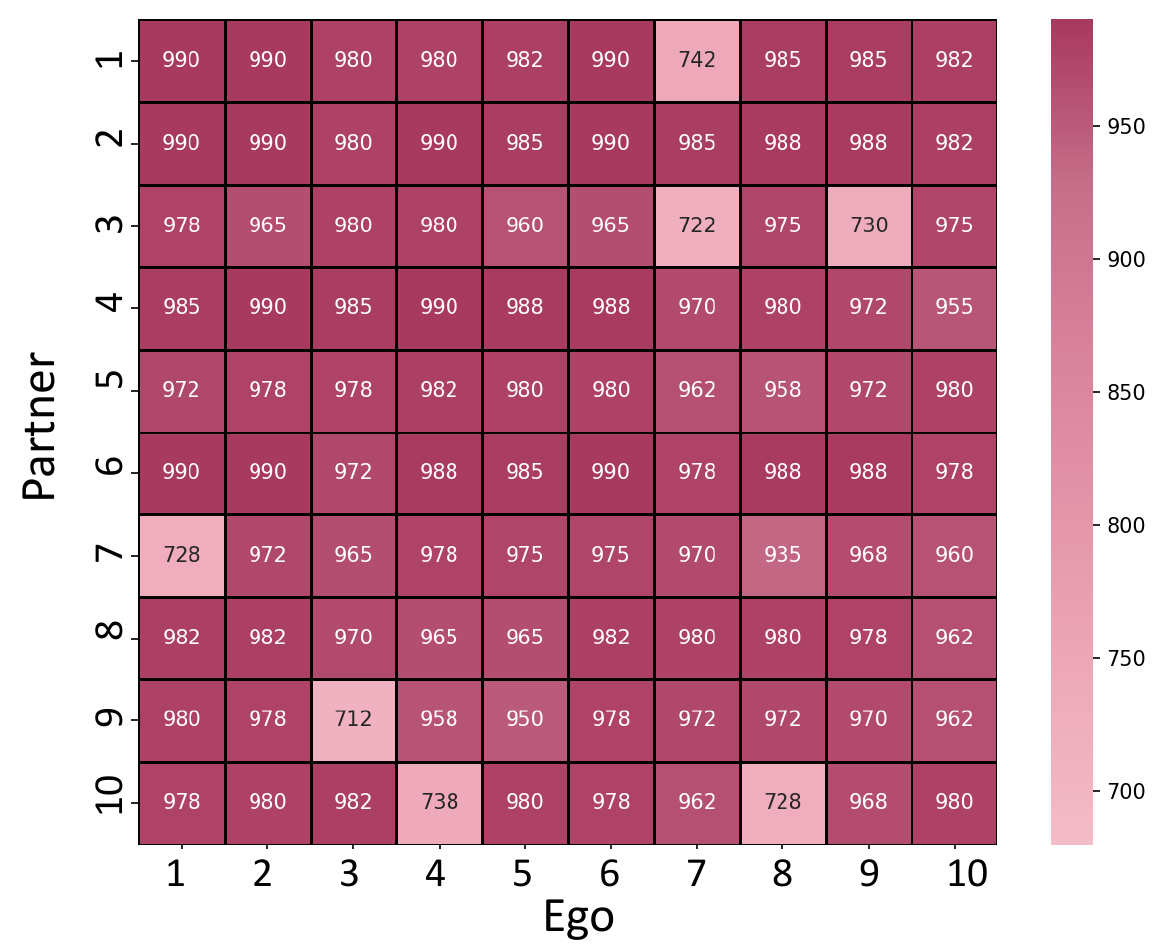}
  \caption{%
    Seed-wise XP heatmaps on the Multi-Destination Spread.
    Rows: ego seeds; Columns: partner seeds.
    Each cell shows the mean episode return for the corresponding ego--partner seed pair.
  }
  \label{fig:heatmap_dual}
\end{figure}

Figure~\ref{fig:heatmap_dual} shows the seed-wise XP performance of SBC on
Multi-Destination Spread by pairing ego policies and partner policies trained
with 10 different random seeds.
The main diagonal entries correspond to SP, where ego and partner policies are
trained with the same seed, and achieve nearly the maximum score.
The off-diagonal entries correspond to XP, where ego and partner policies are
trained with different seeds, and also maintain high rewards.
This indicates that ego policies trained with SBC can flexibly coordinate with
unseen coordination conventions learned from different random seeds.

 \newpage 
 
\begin{figure*}[h!]
  \centering
  \begin{subfigure}[t]{0.49\textwidth}
    \centering
    \includegraphics[width=\linewidth]{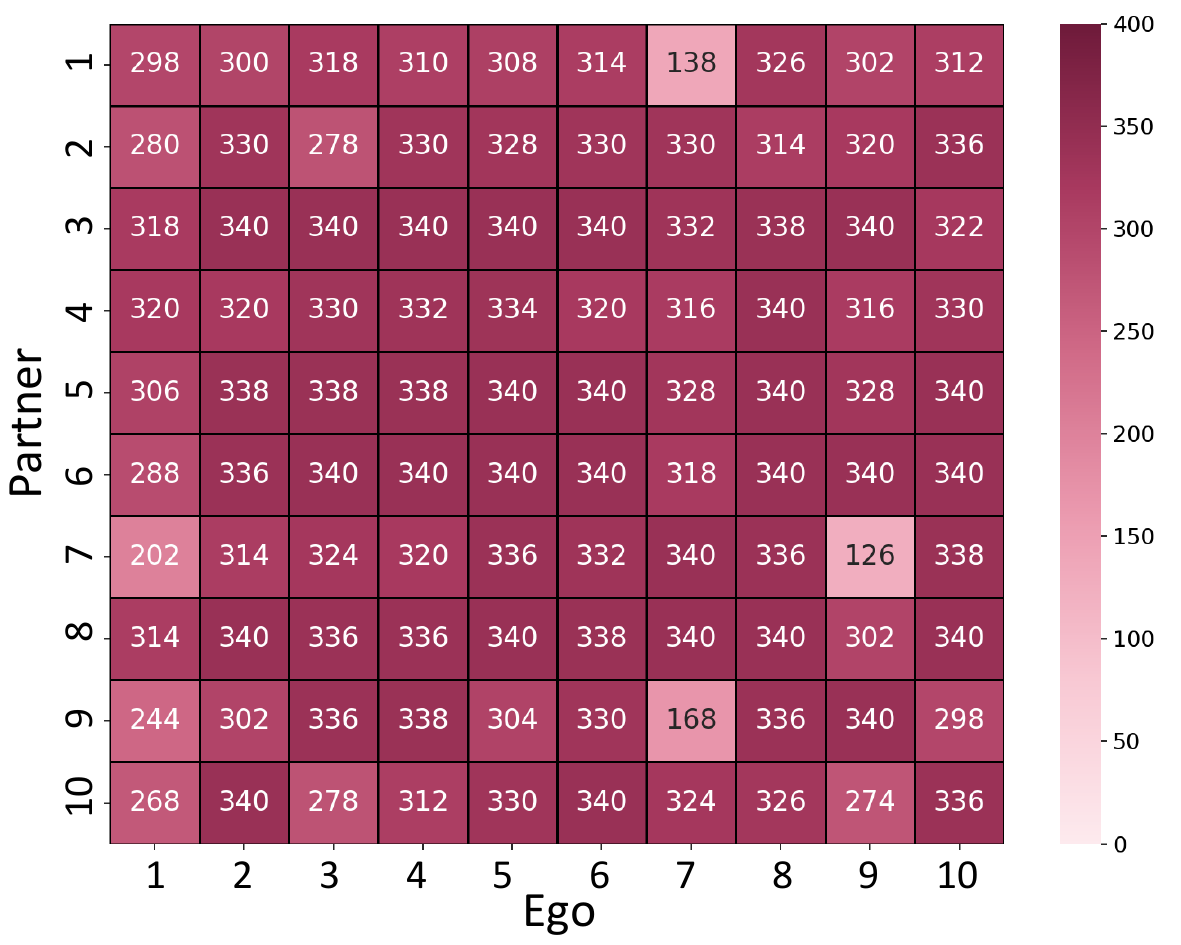}
    \vspace{-0.5cm}
    \caption{\texttt{Coordination Ring}}
    \label{fig:heatmap_ovc1_cr}
  \end{subfigure}
  \hfill
  \begin{subfigure}[t]{0.49\textwidth}
    \centering
    \includegraphics[width=\linewidth]{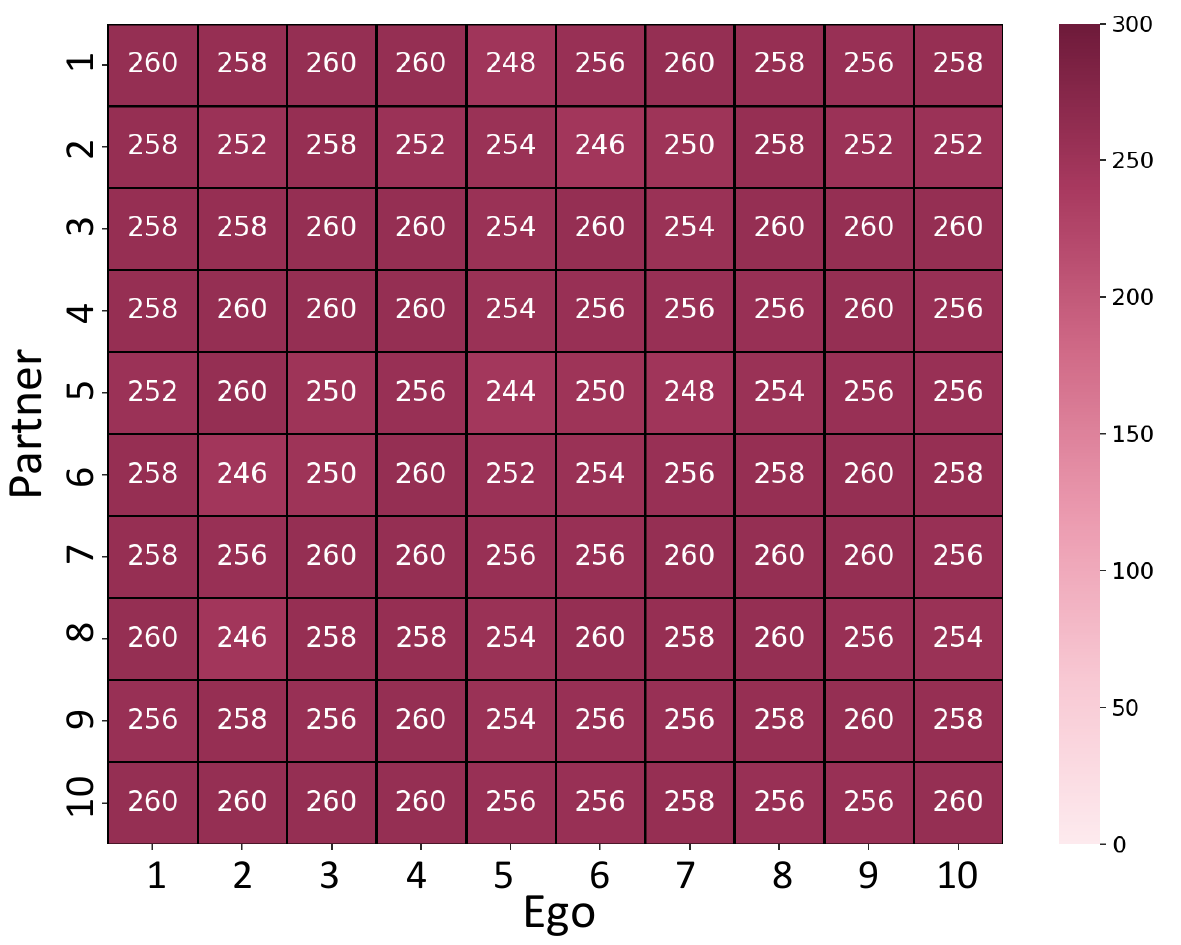}
    \vspace{-0.5cm}
    \caption{\texttt{Cramped Room}}
    \label{fig:heatmap_ovc1_crm}
  \end{subfigure}
  \caption{Seed-wise XP heatmaps on Overcooked v1.}
  \label{fig:heatmap_ovc1}
\end{figure*}

\paragraph{Overcooked v1.} 
\label{app:heatmap_ovc1}
Figure~\ref{fig:heatmap_ovc1} shows seed-wise XP performance on \texttt{Coordination Ring} and \texttt{Cramped Room} by pairing SBC policies trained with different random seeds.
Cells on the main diagonal represent same-seed pairings, while off-diagonal cells represent cross-seed pairings.
In both layouts, cross-seed XP performance is close to the maximum score across nearly all seed pairs.
These results indicate that, by training with diverse locally optimal blocking-aware partners, SBC learns an ego policy that can coordinate well with unseen partners trained from different seeds.

\newpage 

\subsection{Cross-Method Held-out XP}
\label{app:xp_heldout}

We evaluate held-out XP on Multi-Destination Spread and Overcooked v1 to test
whether each ego policy can coordinate with partners trained by different ZSC
methods.
Each cell $(i,j)$ reports the XP score obtained by pairing the ego policy of
method $i$ with partner policies of method $j$, averaged over 10 random seeds.
Same-method pairings are computed via seed-wise cross-play to provide a comparable
reference, while cross-method pairings evaluate generalization to unseen
method-induced partners.
The rightmost \textit{Mean} column reports the row-wise average over 
\emph{cross-method} partner columns only (i.e., excluding the diagonal entry 
$(i,i)$ to isolate generalization to method-induced partners).

\begin{figure}[h!]
  \centering
  \includegraphics[width=0.6\linewidth]{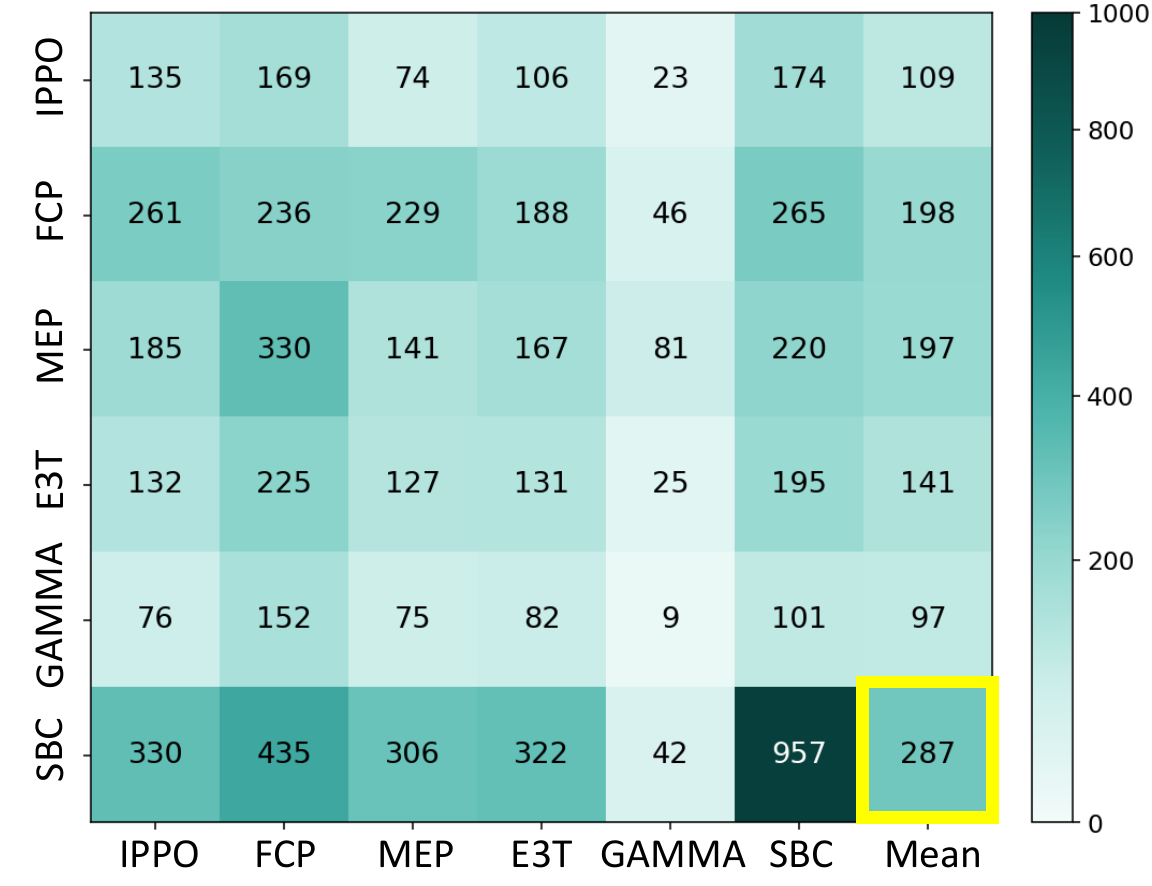}
  \vspace{-0.2cm}
  \caption{%
    Held-out XP heatmap on Multi-Destination Spread.
  }
  \label{fig:xp_heldout_mds}
\end{figure}

\paragraph{Multi-Destination Spread.}
Figure~\ref{fig:xp_heldout_mds} reports cross-method held-out XP performance on
Multi-Destination Spread, where the ego agent is paired with three partners drawn
from another method.
SBC achieves the highest mean held-out XP.
When paired with partners from other methods, SBC often achieves higher scores
than those methods' same-method XP scores, whereas other methods suffer large
drops under cross-method pairing.
The result suggests that blocking-aware partners induce diverse
coordination conventions that cover not only seed-induced conventions but also
conventions learned by other ZSC methods.

\newpage 

\begin{figure*}[h!]
  \centering
  \includegraphics[width=\linewidth]{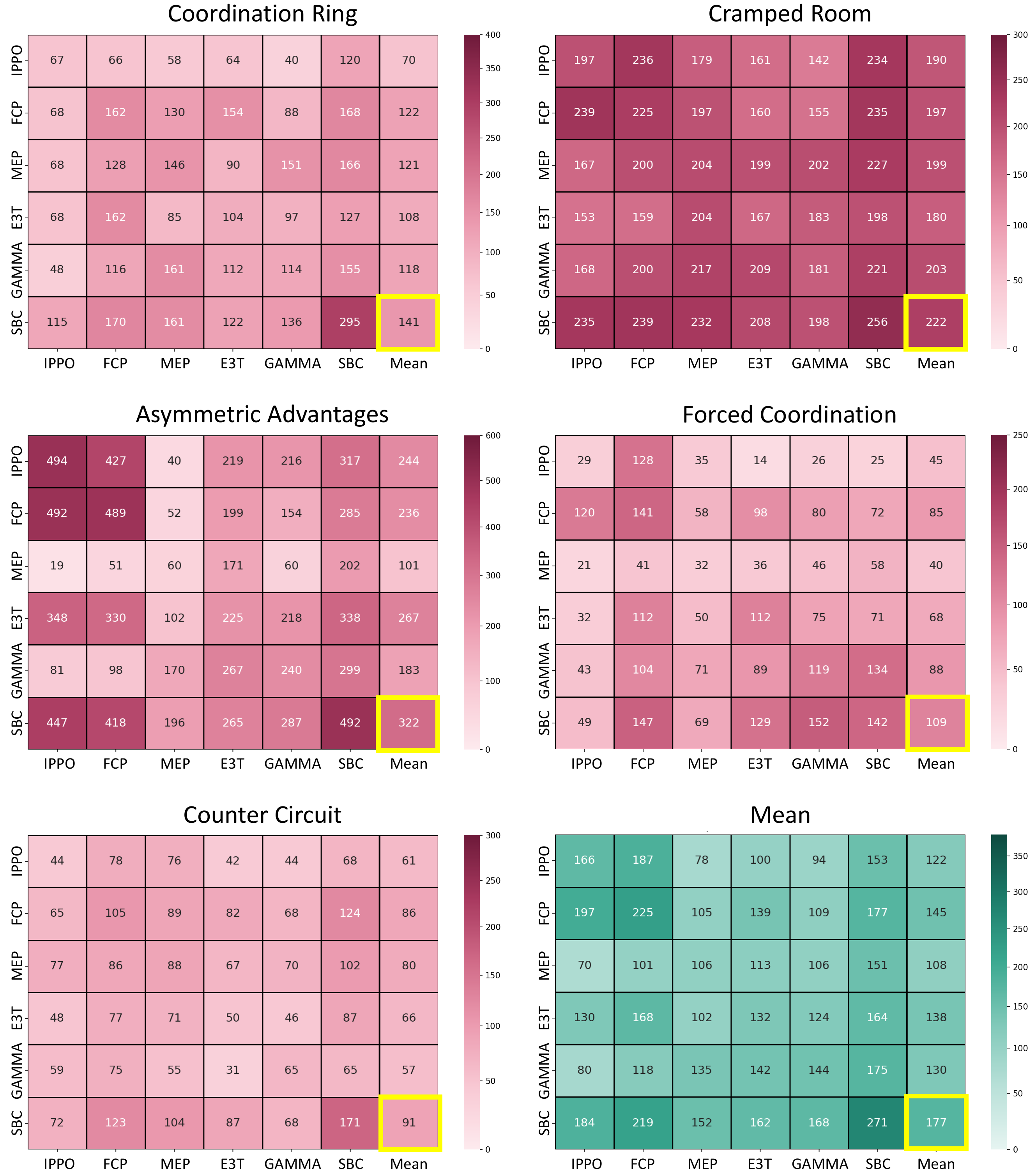}
  \vspace{-0.3cm}
  \caption{%
    Held-out XP  heatmaps on Overcooked v1.
    The bottom-right panel shows the mean across all five layouts.
  }
  \label{fig:xp_heldout_ovc1}
\end{figure*}

\paragraph{Overcooked v1.}
Figure~\ref{fig:xp_heldout_ovc1} reports cross-method held-out XP performance on
all five Overcooked v1 layouts, along with the aggregated mean across layouts in
the bottom-right panel.
Similar to Multi-Destination Spread, most prior methods show substantial
performance drops when paired with partners from different methods compared to
their same-method XP.
In contrast, SBC maintains high held-out XP across partner methods and often
outperforms the same-method XP scores of the corresponding partner methods.
These results show that SBC robustly generalizes to unseen partners induced by
both different random seeds and different ZSC methods.

\newpage 

\subsection{Full Table \& Learning Curves for Overcooked v1}
\label{app:learning_curves}

\begin{figure}[h!]
  \centering
  \includegraphics[width=\linewidth]{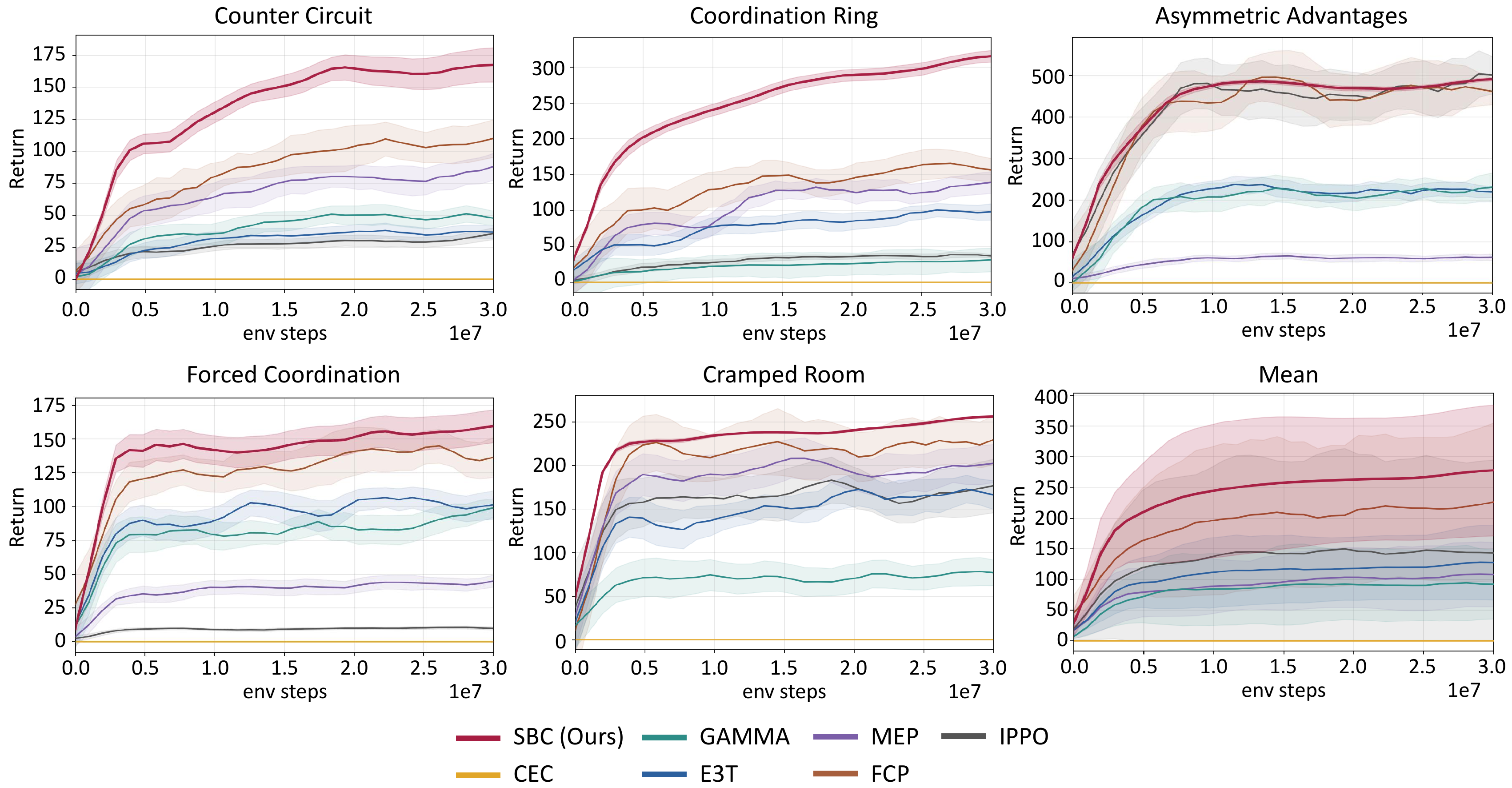}
  \caption{%
    Learning curves on Overcooked v1. The solid lines show the mean cross-play 
    (XP) return averaged over 10 random seeds, whereas the shaded areas indicate 
    the standard deviation.
}
  \label{fig:learning_curves}
\end{figure}

Table~\ref{tab:overcooked_v1_appendix} reports the full episode returns and Figure~\ref{fig:learning_curves} shows the corresponding learning curves on Overcooked v1, evaluated by self-play (SP) and cross-play (XP) scores across five layouts. All values are reported as mean and standard deviation over 10 random seeds. Across all layouts, SBC consistently achieves the highest XP performance while maintaining a small Gap, indicating strong generalization without sacrificing SP performance. In contrast, prior methods often exhibit a large Gap, suggesting overfitting to self-play partners. The learning curves further support these observations as SBC not only converges to higher final returns but also demonstrates faster convergence to the optimal.

\begin{table}[h!]
  \centering
  \small
  \caption{%
   Episode return on Overcooked v1. Values are mean $\pm$ standard deviation across 10 random seeds. \textbf{Bold} indicates the best results within each layout.
  }
  \label{tab:overcooked_v1_appendix}
  \resizebox{\textwidth}{!}{%
  \setlength{\tabcolsep}{4pt}
  \begin{tabular}{l
    r@{$\;\pm\;$}l r@{$\;\pm\;$}l r
    @{\quad}
    r@{$\;\pm\;$}l r@{$\;\pm\;$}l r
    @{\quad}
    r@{$\;\pm\;$}l r@{$\;\pm\;$}l r}
    \toprule
    & \multicolumn{5}{c}{\textbf{Counter Circuit}}
    & \multicolumn{5}{c}{\textbf{Coordination Ring}}
    & \multicolumn{5}{c}{\textbf{Cramped Room}} \\
    \cmidrule(lr){2-6} \cmidrule(lr){7-11} \cmidrule(lr){12-16}
    \textbf{Method}
      & \multicolumn{2}{c}{SP} & \multicolumn{2}{c}{XP} & $\text{Gap}$
      & \multicolumn{2}{c}{SP} & \multicolumn{2}{c}{XP} & $\text{Gap}$
      & \multicolumn{2}{c}{SP} & \multicolumn{2}{c}{XP} & $\text{Gap}$ \\
    \midrule
    IPPO
      & $158$           & $27$           & $\phantom{0}32$ & $12$          & $126$
      & $287$           & $23$           & $\phantom{0}41$ & $11$          & $246$
      & $253$           & $\phantom{0}9$ & $187$           & $36$          & $\phantom{0}66$ \\
    FCP
      & $103$           & $43$           & $107$           & $17$          & $\phantom{00}\mathbf{4}$
      & $159$           & $29$           & $166$           & $11$          & $\phantom{00}7$
      & $224$           & $50$           & $227$           & $19$          & $\phantom{00}3$ \\
    MEP
      & $\phantom{0}87$ & $24$           & $\phantom{0}83$ & $11$          & $\phantom{00}\mathbf{4}$
      & $142$           & $28$           & $143$           & $11$          & $\phantom{00}\mathbf{1}$
      & $203$           & $17$           & $205$           & $\phantom{0}7$ & $\phantom{00}2$ \\
    E3T
      & $121$           & $31$           & $\phantom{0}38$ & $\phantom{0}9$ & $\phantom{0}83$
      & $181$           & $58$           & $\phantom{0}98$ & $24$          & $\phantom{0}83$
      & $234$           & $\phantom{0}6$ & $165$           & $27$          & $\phantom{0}69$ \\
    GAMMA
      & $\phantom{0}66$ & $29$           & $\phantom{0}74$ & $14$          & $\phantom{00}8$
      & $143$           & $64$           & $142$           & $23$          & $\phantom{00}\mathbf{1}$
      & $181$           & $24$           & $182$           & $\phantom{0}7$ & $\phantom{00}\mathbf{1}$ \\
    CEC
      & $\phantom{0}82$ & $31$           & $\phantom{0}52$ & $12$          & $\phantom{0}30$
      & $203$           & $40$           & $175$           & $16$          & $\phantom{0}28$
      & $221$           & $21$           & $219$           & $19$          & $\phantom{00}2$ \\
    \midrule
    \textbf{SBC}
      & $\mathbf{246}$  & $\mathbf{38}$  & $\mathbf{168}$  & $\mathbf{32}$ & $\phantom{0}78$
      & $\mathbf{333}$  & $\mathbf{13}$  & $\mathbf{316}$  & $\mathbf{18}$ & $\phantom{0}17$
      & $\mathbf{257}$  & $\phantom{0}\mathbf{5}$ & $\mathbf{256}$ & $\phantom{0}\mathbf{1}$ & $\phantom{00}\mathbf{1}$ \\
    \bottomrule
  \end{tabular}%
  }
  \vspace{0.4cm}
  \resizebox{0.7\textwidth}{!}{%
  \setlength{\tabcolsep}{4pt}
  \begin{tabular}{l
    r@{$\;\pm\;$}l r@{$\;\pm\;$}l r
    @{\quad}
    r@{$\;\pm\;$}l r@{$\;\pm\;$}l r}
    \toprule
    & \multicolumn{5}{c}{\textbf{Forced Coordination}}
    & \multicolumn{5}{c}{\textbf{Asymmetric Advantages}} \\
    \cmidrule(lr){2-6} \cmidrule(lr){7-11}
    \textbf{Method}
      & \multicolumn{2}{c}{SP} & \multicolumn{2}{c}{XP} & $\text{Gap}$
      & \multicolumn{2}{c}{SP} & \multicolumn{2}{c}{XP} & $\text{Gap}$ \\
    \midrule
    IPPO
      & $\mathbf{200}$  & $\mathbf{29}$  & $\phantom{0}10$ & $11$          & $190$
      & $\mathbf{500}$  & $\phantom{0}\mathbf{1}$  & $\mathbf{499}$ & $\phantom{0}\mathbf{9}$ & $\phantom{0}1$ \\
    FCP
      & $128$           & $37$           & $140$           & $\phantom{0}9$ & $\phantom{0}12$
      & $498$           & $\phantom{0}3$ & $479$           & $25$          & $19$ \\
    MEP
      & $\phantom{0}38$ & $21$           & $\phantom{0}43$ & $10$          & $\phantom{00}\mathbf{5}$
      & $\phantom{0}40$ & $74$           & $\phantom{0}64$ & $50$          & $24$ \\
    E3T
      & $176$           & $10$           & $105$           & $22$          & $\phantom{0}71$
      & $229$           & $13$           & $226$           & $\phantom{0}4$ & $\phantom{0}3$ \\
    GAMMA
      & $108$           & $35$           & $129$           & $11$          & $\phantom{0}21$
      & $240$           & $\phantom{0}1$ & $240$           & $\phantom{0}1$ & $\phantom{0}\mathbf{0}$ \\
    CEC
      & $112$           & $26$           & $\phantom{0}81$ & $13$          & $\phantom{0}31$
      & $283$           & $18$           & $251$           & $17$          & $32$ \\
    \midrule
    \textbf{SBC}
      & $193$           & $10$           & $\mathbf{160}$  & $\mathbf{23}$ & $\phantom{0}33$
      & $\mathbf{500}$  & $\mathbf{21}$  & $493$           & $10$          & $\phantom{0}7$ \\
    \bottomrule
  \end{tabular}%
  }
\end{table}
 
\newpage 

\renewcommand{\theequation}{E.\arabic{equation}}
\setcounter{equation}{0}
\renewcommand{\thefigure}{E.\arabic{figure}}
\setcounter{figure}{0}
\renewcommand{\thetable}{E.\arabic{table}}
\setcounter{table}{0}
\section{Additional Analysis}
\label{app:analysis}
This section provides additional analyses of SBC. While Appendix~\ref{app:traj_full} analyzes how SBC generalizes well to unseen partners
through trajectory analysis, Appendices~\ref{app:component_full} and \ref{app:hyperparam_full} extend the component evaluations and ablations on hyperparameters to additional Overcooked v1 layouts. Finally, Section~\ref{app:compute} compares the computational cost of SBC against the considered baselines.

\subsection{Human--AI Behavior Analysis}
\label{app:traj_full}
\begin{figure}[h!]
  \centering
  \includegraphics[width=0.55\linewidth]{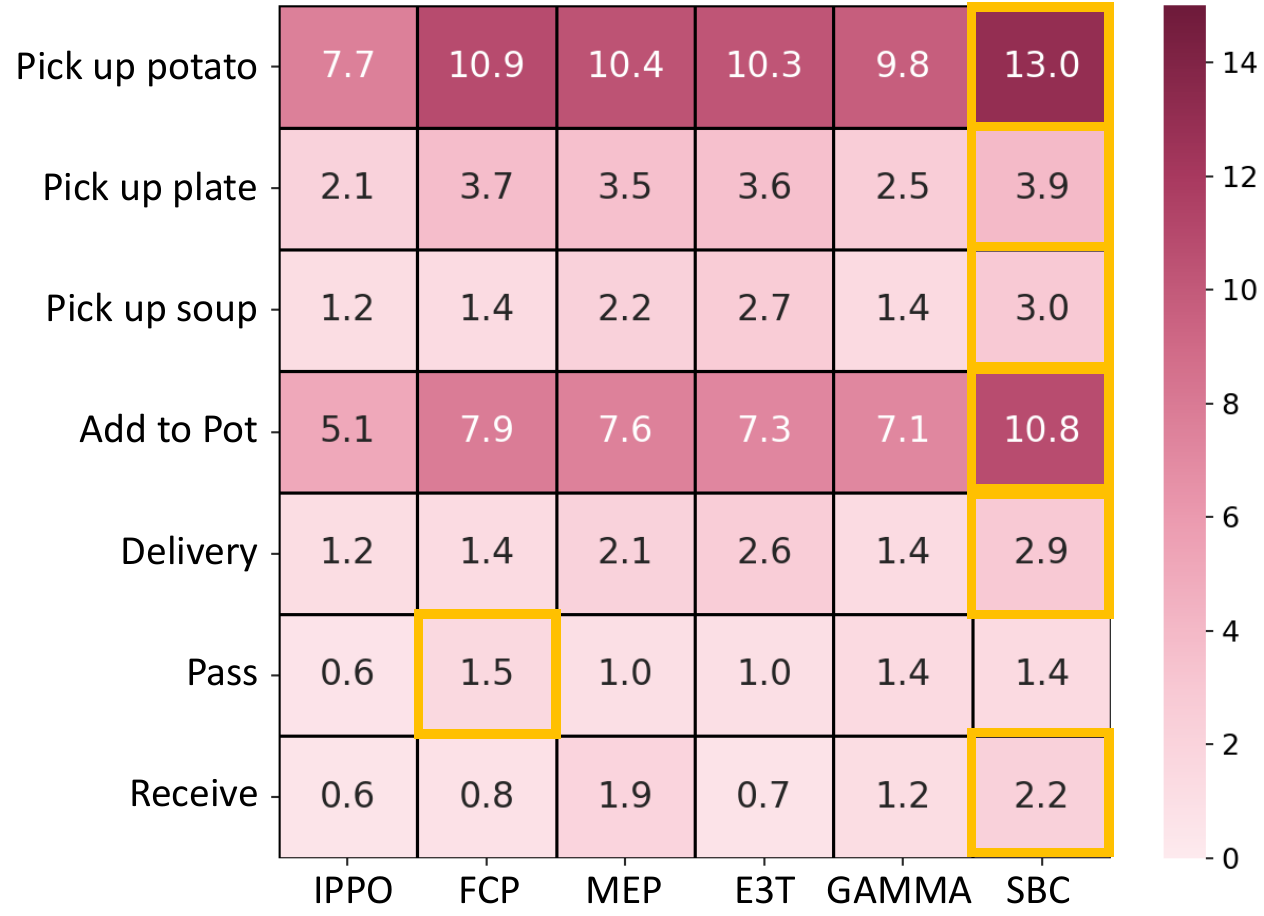}
  \caption{%
    Per-episode mean counts of seven behavioral primitives, averaged across all
    five Overcooked v1 layouts. Darker cells indicate higher counts.
  }
  \label{fig:traj_convergence_full}
\end{figure}

\begin{figure}[h!]
  \centering
  \includegraphics[width=1.0\linewidth]{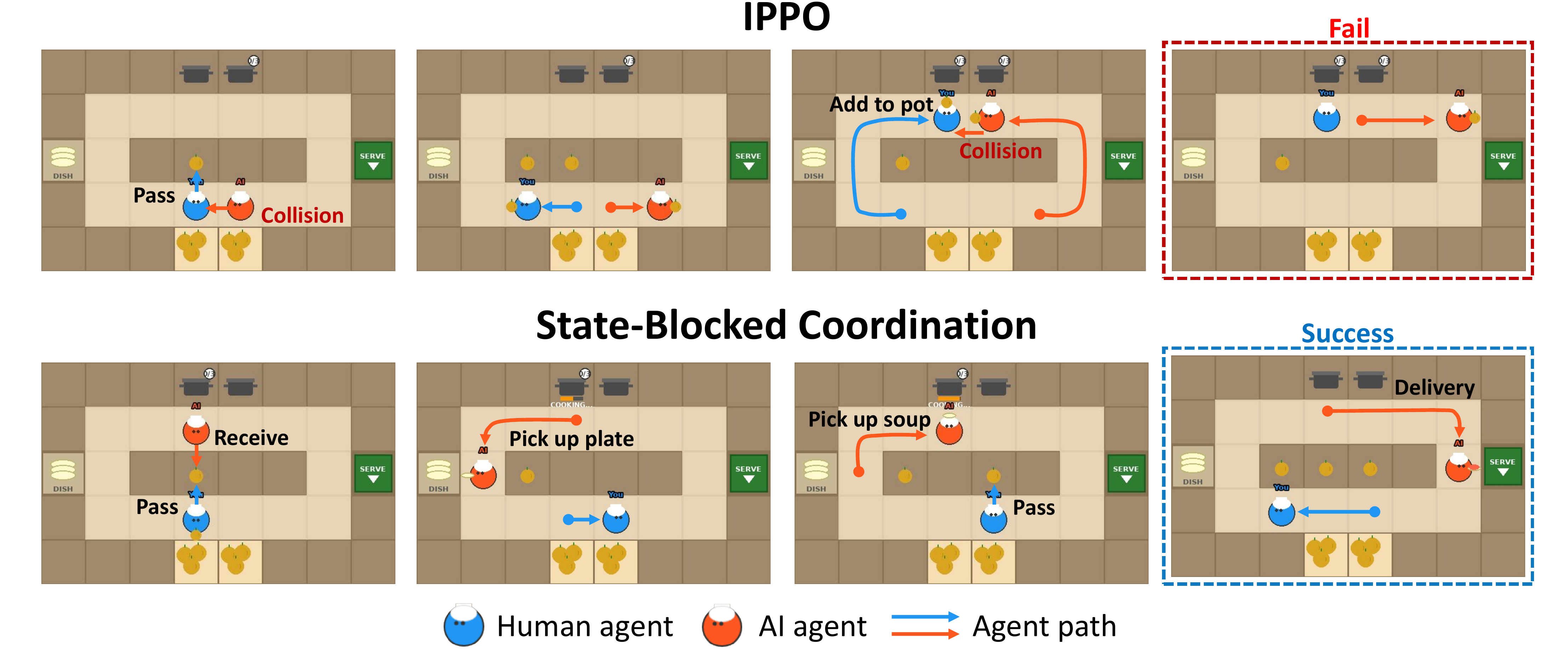}
  \caption{%
    Visualization of IPPO and SBC behaviors when the human agent passes an ingredient
    on \texttt{Counter Circuit} in Overcooked v1.
  }
  \label{fig:behavioral_full}
\end{figure}

This section analyzes human--AI trajectories to explain why SBC coordinates well
with diverse unseen human partners.
Figure~\ref{fig:traj_convergence_full} reports the per-episode mean counts of
seven behavioral primitives, averaged across all five Overcooked v1 layouts and
human--AI trajectories.
Prior baselines mainly show low-level behaviors such as simple pick-up actions,
with fewer deliveries and rare passing or receiving.
In contrast, SBC exhibits substantially higher counts of cooperative behaviors,
including passing, receiving, and frequent deliveries.
Figure~\ref{fig:behavioral_full} provides a qualitative example.
When the human agent passes an ingredient, IPPO fails to adapt, collides with the
human agent, and continues acting independently.
By contrast, SBC receives the ingredient and completes the delivery.
Together, these results suggest that SBC learns a broader set of human-compatible
coordination patterns through exposure to blocking-aware partners, enabling the
ego policy to adapt to unseen human behaviors.

\newpage

\subsection{Component ablation across all layouts}
\label{app:component_full}
Table~\ref{tab:component_ablation_appendix} extends the component ablation to all five Overcooked v1 layouts. According to the results, removing the blocking-aware partner causes the largest average drop, highlighting its central role in enabling robust coordination under diverse partner behaviors. Removing the SP partner also leads to substantial degradation, particularly in coordination-intensive layouts such as Counter Circuit and Coordination Ring, where learning stable conventions is essential. Furthermore, Uniform sampling and strict penalties consistently hurt performance, indicating the importance of allowing flexible adaptation. On the other hand, Action prediction leads to smaller gains, suggesting it provides complementary improvements in modeling partner behavior.
Overall, while the magnitude of impact differs across components and layouts, all components contribute to performance, and the full SBC design is necessary to achieve strong and consistent coordination across diverse settings.

\begin{table}[H]
  \centering
  \small
  \caption{Component ablation across Overcooked v1 layouts (XP score). \textbf{Mean} is the average across layouts. \textbf{Bold} marks the best per column.}
  \label{tab:component_ablation_appendix}
  \resizebox{\textwidth}{!}{%
  \setlength{\tabcolsep}{4pt}
  \begin{tabular}{l
    r@{$\;\pm\;$}l
    r@{$\;\pm\;$}l
    r@{$\;\pm\;$}l
    r@{$\;\pm\;$}l
    r@{$\;\pm\;$}l
    c}
    \toprule
    \textbf{Variant}
      & \multicolumn{2}{c}{\textbf{Counter Circuit}}
      & \multicolumn{2}{c}{\textbf{Coord. Ring}}
      & \multicolumn{2}{c}{\textbf{Cramped Room}}
      & \multicolumn{2}{c}{\textbf{Forced Coord.}}
      & \multicolumn{2}{c}{\textbf{Asymm. Adv.}}
      & \textbf{Mean} \\
    \midrule
    \textbf{SBC (Ours)}
      & $\mathbf{168}$ & $\mathbf{32.0}$
      & $\mathbf{316}$ & $\mathbf{18.0}$
      & $\mathbf{256}$ & $\phantom{0}\mathbf{1.0}$
      & $\mathbf{160}$ & $\mathbf{23.0}$
      & $\mathbf{493}$ & $\mathbf{10.0}$
      & $\mathbf{278.6}$ \\
    \midrule
    \;\;w/ uniform sampling
      & $119$          & $30.4$
      & $257$          & $23.8$
      & $247$          & $\phantom{0}3.4$
      & $115$          & $29.6$
      & $459$          & $17.8$
      & $239.4$ \\
    \;\;w/o action prediction
      & $142$          & $28.6$
      & $193$          & $94.8$
      & $244$          & $\phantom{0}4.6$
      & $103$          & $44.9$
      & $448$          & $48.0$
      & $226.0$ \\
    \;\;w/ strict penalty
      & $123$          & $25.3$
      & $208$          & $38.7$
      & $242$          & $\phantom{0}4.9$
      & $\phantom{0}74$ & $30.2$
      & $468$          & $\phantom{0}8.3$
      & $223.0$ \\
    \;\;w/o SP partner
      & $\phantom{0}69$ & $19.7$
      & $\phantom{0}40$ & $35.4$
      & $\phantom{0}70$ & $22.8$
      & $\phantom{0}76$ & $21.4$
      & $469$          & $13.7$
      & $144.8$ \\
    \;\;w/o blocking-aware partner
      & $\phantom{0}38$ & $\phantom{0}8.5$
      & $\phantom{0}98$ & $17.0$
      & $165$          & $14.2$
      & $105$          & $11.3$
      & $226$          & $\phantom{0}2.8$
      & $126.4$ \\
    \bottomrule
  \end{tabular}%
  }
\end{table}

 \newpage 
 
\subsection{Hyperparameter ablation on three additional layouts}
\label{app:hyperparam_full}
In addition to \texttt{Counter Circuit} reported in Section~\ref{sec:ablation}, we analyze the effect of the penalty coefficient $\alpha$ and the maximum size of the penalty-state set $K$ on three additional Overcooked v1 layouts: \texttt{Coordination Ring}, \texttt{Cramped Room}, and \texttt{Forced Coordination}.
Most importantly, we observe that across all layouts, removing the penalty signal (i.e., $\alpha = 0$ or $K = 0$) consistently yields the worst cross-play (XP) performance, highlighting that penalty-based guidance is essential for learning robust coordination.

As $\alpha$ increases from zero, XP performance improves substantially, but overly large values eventually degrade performance, revealing a trade-off between coordination and over-penalizing useful behaviors. A similar pattern is observed when varying $K$. Increasing $K$ from zero significantly improves XP, confirming the importance of capturing coordination failures, while larger values lead to diminishing returns or slight degradation, suggesting that excessively broad penalty-state sets introduce noise. Overall, these results demonstrate that penalty signals are critical for coordination, and that choosing appropriate $\alpha$ and $K$ can provide informative guidance without over-constraining the policy.
\begin{figure}[H]
  \centering
  \includegraphics[width=0.9\linewidth]{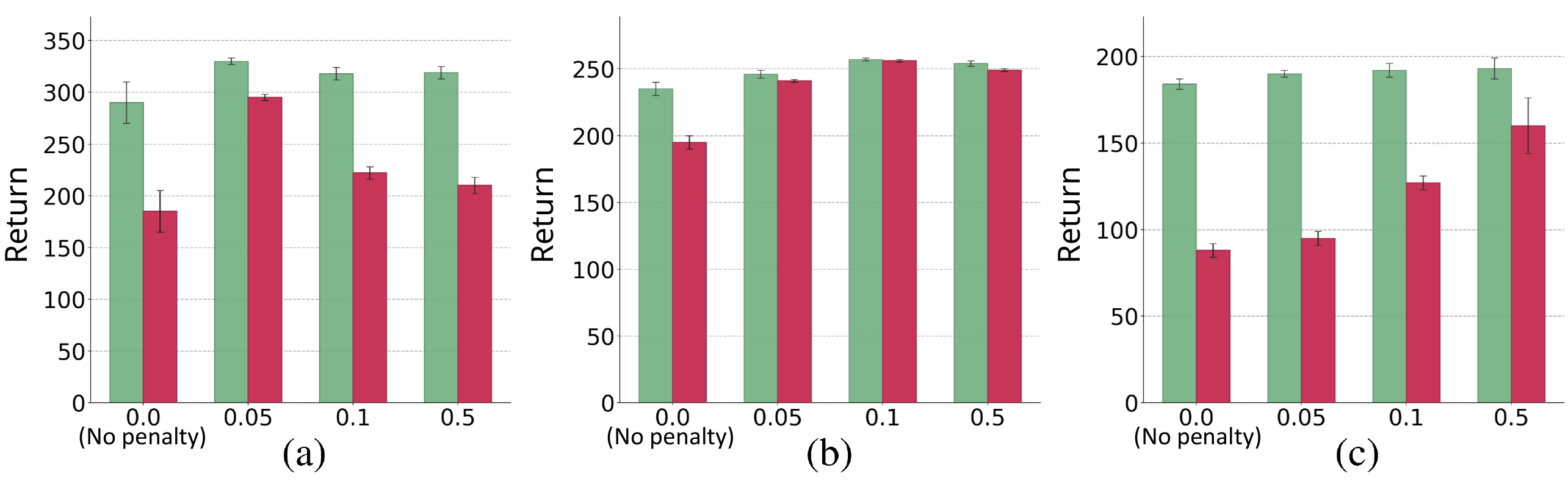}
  \caption{%
    Penalty coefficient $\alpha$ sweep.
    \textbf{(a)} \texttt{Coordination Ring},
    \textbf{(b)} \texttt{Cramped Room},
    \textbf{(c)} \texttt{Forced Coordination}.
    Green: self-play; red: cross-play.
  }
  \label{fig:hp_ablation_alpha}
\end{figure}

\begin{figure}[H]
  \centering
  \includegraphics[width=0.9\linewidth]{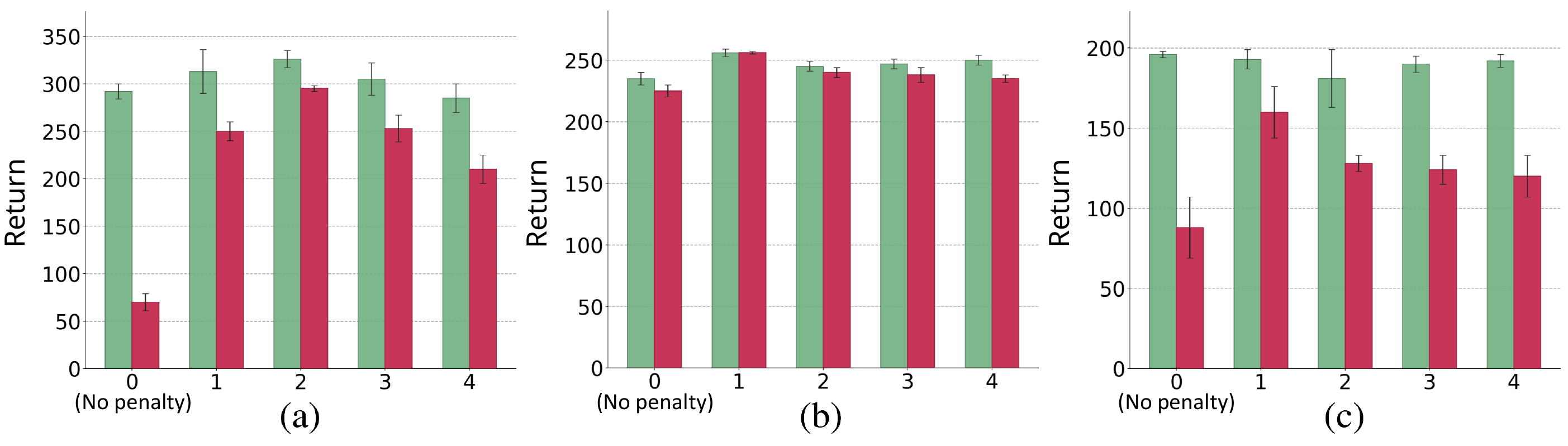}
  \caption{%
    Maximum size of the penalty-state set $K$ sweep.
    \textbf{(a)} \texttt{Coordination Ring},
    \textbf{(b)} \texttt{Cramped Room},
    \textbf{(c)} \texttt{Forced Coordination}.
  }
  \label{fig:hp_ablation_k}
\end{figure}

\newpage 

\newpage

\subsection{Computational Cost}
\label{app:compute}
Table~\ref{tab:compute} reports the total training time required to reach 30M environment steps across Overcooked layouts, averaged over 10 seeds. For SBC, the 30M budget refers to the combined total environment interactions across both the normal policy and the blocking-aware policy rollouts, ensuring sample-budget parity with all baselines. For FCP, the reported time includes both the construction of the diversity pool and the subsequent pool-based training. For GAMMA, we report the time required to reach 100M steps, which is necessary to obtain reasonable performance. For CEC, we report the total time for a single unified training run over all layouts (3000M steps), as it leverages cross-layout experience rather than training separate policies per layout.

Overall, SBC introduces a moderate computational overhead compared to lightweight baselines such as IPPO and E3T. On average, SBC requires 2\,h\,27\,m of training, compared to 1\,h\,33\,m for IPPO and 1\,h\,40\,m for E3T, corresponding to an increase of approximately  +58\% over IPPO. However, this additional cost is modest relative to the substantial gains in coordination performance, especially since SBC remains significantly more efficient than methods with heavier training procedures, such as GAMMA, which is approximately $2.7$ times slower than SBC. These results demonstrate that SBC provides an efficient approach to learning robust coordination, achieving strong performance improvements with a reasonable increase in training time.

\begin{table}[h!]
  \centering
  \caption{%
    Total training time comparison across Overcooked layouts
    (\,h:\,hour,\; m:\,minute).
    Times are averaged over $10$ seeds.
  }
  \label{tab:compute}
  \resizebox{\textwidth}{!}{%
  \begin{tabular}{l ccccc c}
    \toprule
    \multirow{2}{*}{\textbf{Method}}
      & \textbf{Cramped}  & \textbf{Asymmetric} & \textbf{Coord.}      & \textbf{Forced}      & \textbf{Counter}
      & \multirow{2}{*}{\textbf{Mean}} \\
      & \textbf{Room}     & \textbf{Advantages} & \textbf{Ring}         & \textbf{Coord.}      & \textbf{Circuit}
      & \\
    \midrule
    IPPO
      & 1\,h\;17\,m & 1\,h\;58\,m & 1\,h\;24\,m & 1\,h\;23\,m & 1\,h\;41\,m
      & 1\,h\;33\,m \\
    FCP
      & 2\,h\;49\,m & 3\,h\;43\,m & 2\,h\;52\,m & 2\,h\;53\,m & 3\,h\;45\,m
      & 3\,h\;12\,m \\
    MEP
      & 1\,h\;52\,m & 2\,h\;54\,m & 2\,h\;04\,m & 3\,h\;08\,m & 3\,h\;32\,m
      & 2\,h\;42\,m \\
    E3T
      & 1\,h\;06\,m & 1\,h\;50\,m & 1\,h\;15\,m & 1\,h\;28\,m & 2\,h\;41\,m
      & 1\,h\;40\,m \\
    GAMMA
      & 5\,h\;28\,m & 8\,h\;46\,m & 5\,h\;50\,m & 6\,h\;17\,m & 6\,h\;27\,m
      & 6\,h\;33\,m \\
    CEC 
      & \multicolumn{5}{c}{44\,h\;15\,m\;\textit{(single unified training)}} & 8\,h\;51\,m \\
    \midrule
    \textbf{SBC (Ours)}
      & 2\,h\;20\,m & 2\,h\;34\,m & 2\,h\;29\,m & 2\,h\;21\,m & 2\,h\;30\,m
      & 2\,h\;27\,m \\
    \bottomrule
  \end{tabular}%
  }
\end{table}

\section{Broader Impacts}
\label{app:broader_impacts}

This work studies zero-shot coordination, aiming to train agents that
cooperate effectively with unseen partners, including humans. Such methods may
contribute to broader research on human coordination, human--AI collaboration,
and AI--AI coordination by enabling agents to adapt to partners with different
conventions, preferences, and behaviors. This has potential applications in
collaborative robotics, shared-autonomy systems, mixed human--robot teams, and
cooperative multi-agent systems where independently trained agents must
coordinate without prior joint training. More broadly, improving zero-shot
coordination can support the development of cooperative AI systems that interact
more flexibly with both human and artificial partners. Since our method is trained and evaluated in simulation, real-world deployment
would require validation for sim-to-real robustness and safety.

\clearpage

\section*{NeurIPS Paper Checklist}

\begin{enumerate}

\item {\bf Claims}
    \item[] Question: Do the main claims made in the abstract and introduction accurately reflect the paper's contributions and scope?
    \item[] Answer: \answerYes{}
    \item[] Justification: The abstract and introduction clearly state our main contributions, including the SBC framework with value-guided penalty-state scheduling, and our evaluation on Multi-Destination Spread and Overcooked v1, including human--AI coordination, which are supported by the experimental results in Section~\ref{sec:experiments} and Section~\ref{sec:human_eval}.

\item {\bf Limitations}
    \item[] Question: Does the paper discuss the limitations of the work performed by the authors?
    \item[] Answer: \answerYes{}
    \item[] Justification: We discuss the limitations of our work in Section~\ref{sec:limitations}, including the additional hyperparameters introduced by SBC, scaling behavior to settings with more agents, and the computational overhead of value-gap estimation.

\item {\bf Theory assumptions and proofs}
    \item[] Question: For each theoretical result, does the paper provide the full set of assumptions and a complete (and correct) proof?
    \item[] Answer: \answerNA{}
    \item[] Justification: The paper does not include theoretical results; our contributions are empirical.

    \item {\bf Experimental result reproducibility}
    \item[] Question: Does the paper fully disclose all the information needed to reproduce the main experimental results of the paper to the extent that it affects the main claims and/or conclusions of the paper (regardless of whether the code and data are provided or not)?
    \item[] Answer: \answerYes{}
    \item[] Justification: We provide the full SBC algorithm in Section~\ref{sec:train} (Algorithm~\ref{alg:sbc}), implementation details in Appendix~\ref{app:impl}, environment details in Appendix~\ref{app:envs}, and full hyperparameters and evaluation protocols in Appendix~\ref{app:experimental}.

\item {\bf Open access to data and code}
    \item[] Question: Does the paper provide open access to the data and code, with sufficient instructions to faithfully reproduce the main experimental results, as described in supplemental material?
    \item[] Answer: \answerYes{}
    \item[] Justification: We provide an anonymized code package as supplementary material, including instructions to reproduce our main experimental results, with full release planned upon publication. The Overcooked v1 environment and human demonstration data.

\item {\bf Experimental setting/details}
    \item[] Question: Does the paper specify all the training and test details (e.g., data splits, hyperparameters, how they were chosen, type of optimizer) necessary to understand the results?
    \item[] Answer: \answerYes{}
    \item[] Justification: Training and evaluation details, including hyperparameters (Appendix~\ref{app:hyperparameters}), optimizer choices, the number of random seeds (10 per method), episode lengths, and evaluation protocols (Appendix~\ref{app:zsc_eval}), are reported in Section~\ref{sec:experiments} and Appendix~\ref{app:experimental}.

\item {\bf Experiment statistical significance}
    \item[] Question: Does the paper report error bars suitably and correctly defined or other appropriate information about the statistical significance of the experiments?
    \item[] Answer: \answerYes{}
    \item[] Justification: All quantitative results are aggregated over 10 random seeds per method and reported as mean $\pm$ standard deviation, with variability arising from random initialization and seed-induced training stochasticity. For real-human evaluation, we additionally conduct two-sided Wilcoxon signed-rank tests ($p < 0.01$) to assess statistical significance under the within-subjects experimental design, as reported in Section~\ref{sec:human_eval} and Appendix~\ref{app:zsc_eval}.

\item {\bf Experiments compute resources}
    \item[] Question: For each experiment, does the paper provide sufficient information on the computer resources (type of compute workers, memory, time of execution) needed to reproduce the experiments?
    \item[] Answer: \answerYes{}
    \item[] Justification: Compute resources used for training and evaluation are described in Appendix~\ref{app:compute}.
    
\item {\bf Code of ethics}
    \item[] Question: Does the research conducted in the paper conform, in every respect, with the NeurIPS Code of Ethics \url{https://neurips.cc/public/EthicsGuidelines}?
    \item[] Answer: \answerYes{}
    \item[] Justification: All authors have reviewed the NeurIPS Code of Ethics, and the research conforms to it in every respect.

\item {\bf Broader impacts}
    \item[] Question: Does the paper discuss both potential positive societal impacts and negative societal impacts of the work performed?
    \item[] Answer: \answerYes{}
    \item[] Justification: We discuss potential positive and negative societal impacts of our work in Appendix~\ref{app:broader_impacts}.
    
\item {\bf Safeguards}
    \item[] Question: Does the paper describe safeguards that have been put in place for responsible release of data or models that have a high risk for misuse (e.g., pre-trained language models, image generators, or scraped datasets)?
    \item[] Answer: \answerNA{}
    \item[] Justification: Our work does not release pretrained models or datasets that pose a high risk of misuse; the trained policies operate only within benchmark cooperative game simulators (Multi-Destination Spread and Overcooked v1).

\item {\bf Licenses for existing assets}
    \item[] Question: Are the creators or original owners of assets (e.g., code, data, models), used in the paper, properly credited and are the license and terms of use explicitly mentioned and properly respected?
    \item[] Answer: \answerYes{}
    \item[] Justification: We use the open-source Overcooked-AI codebase~\citep{carroll2019utility}, including its human demonstration data and behavior cloning model, with proper citation. Our implementation is also built on top of JaxMARL~\citep{flair2024jaxmarl} as the core training infrastructure. Baseline implementations from open-source releases are also cited in Appendix~\ref{app:other_baselines}.

\item {\bf New assets}
    \item[] Question: Are new assets introduced in the paper well documented and is the documentation provided alongside the assets?
    \item[] Answer: \answerYes{}
    \item[] Justification: We will release our code and the custom web interface used for real-human evaluation, with documentation describing training, evaluation, and study procedures, alongside the anonymized code release.

\item {\bf Crowdsourcing and research with human subjects}
    \item[] Question: For crowdsourcing experiments and research with human subjects, does the paper include the full text of instructions given to participants and screenshots, if applicable, as well as details about compensation (if any)? 
    \item[] Answer: \answerYes{}
    \item[] Justification: We describe the human study protocol in Appendix~\ref{app:zsc_eval}, including the web interface and pre-session instructions shown to participants (Figure~\ref{fig:appendix_interface}). All participants provided informed consent, and were fairly compensated for their time at a rate at or above the local minimum wage in the country where the data was collected, in accordance with the NeurIPS Code of Ethics.

\item {\bf Institutional review board (IRB) approvals or equivalent for research with human subjects}
    \item[] Question: Does the paper describe potential risks incurred by study participants, whether such risks were disclosed to the subjects, and whether Institutional Review Board (IRB) approvals (or an equivalent approval/review based on the requirements of your country or institution) were obtained?
    \item[] Answer: \answerYes{}
    \item[] Justification: Our human study follows the IRB protocol approved by our institution, as stated in Appendix~\ref{app:zsc_eval}. Potential risks were minimal and disclosed to participants through the consent form.

\item {\bf Declaration of LLM usage}
    \item[] Question: Does the paper describe the usage of LLMs if it is an important, original, or non-standard component of the core methods in this research? Note that if the LLM is used only for writing, editing, or formatting purposes and does \emph{not} impact the core methodology, scientific rigor, or originality of the research, declaration is not required.
    \item[] Answer: \answerNA{}
    \item[] Justification: LLMs were used only for writing and editing assistance and did not contribute to the core methodology, scientific rigor, or originality of this research.

\end{enumerate}

\end{document}